\documentclass[twocolumn,floatfix,amssymb,aps,superscriptaddress,letterpager]{revtex4-2}
\usepackage[colorlinks=true,urlcolor=blue,citecolor=blue,linkcolor=blue]{hyperref}
\usepackage{graphicx}
\usepackage{color}
\usepackage{epsfig}
\usepackage{amsmath}
\usepackage{amssymb}
\usepackage{mathrsfs}
\usepackage{bm}
\usepackage{subfigure}
\usepackage{float}
\usepackage{url}
\usepackage{enumerate}
\usepackage{braket}
\usepackage{comment}
\usepackage{multirow}
\usepackage{CJK}
\usepackage{indentfirst}
\usepackage{amsmath}
\usepackage{cases}
\usepackage{hyperref}
\definecolor{violet}{rgb}{0.56,0.0,1.0}
\def\tbib#1{{\bf{#1}}}

\def\lbra#1{\left(#1\right)}

\def\tit#1{{\textit{#1}}}

\begin{document}

\hsize\textwidth\columnwidth\hsize\csname@twocolumnfalse\endcsname

\title{Exploring explicit coarse-grained structure in artificial neural networks}

\author{Xi-Ci Yang}
\affiliation{College of Power and Energy Engineering, Harbin Engineering University, Harbin 150001, China}
\author{Z. Y. Xie}
\email[]{qingtaoxie@ruc.edu.cn}
\affiliation{Department of Physics,  Renmin University of China,  Beijing 100872,  China}
\author{Xiao-Tao Yang}
\email[]{yangxiaotao@hrbeu.edu.cn}
\affiliation{College of Power and Energy Engineering, Harbin Engineering University, Harbin 150001, China}

\begin{abstract}
We propose to employ the hierarchical coarse-grained structure in the artificial neural networks explicitly to improve the interpretability without degrading performance. The idea has been applied in two situations. One is a neural network called TaylorNet, which aims to approximate the general mapping from input data to output result in terms of Taylor series directly, without resorting to any magic nonlinear activations. The other is a new setup for data distillation, which can perform multi-level abstraction of the input dataset and generate new data that possesses the relevant features of the original dataset and can be used as references for classification. In both cases, the coarse-grained structure plays an important role in simplifying the network and improving both the interpretability and efficiency. The validity has been demonstrated on MNIST and CIFAR-10 datasets. Further improvement and some open questions related are also discussed.
\end{abstract}

\pacs{}
\maketitle
\section{Introduction}
In the past decade, machine learning has drawn great attention from almost all natural science and engineering communities, such as mathematics \cite{GXF-arXiv2018, GXF-CAGD2019, AF-Nature2022}, physics \cite{GX-NC2017, Melko-NP2017, ZP-PRL2019, Cirac-RMP2019, Edwin-Rev2021, Owen-JCP2021, Millis-PRL2022}, biology \cite{Helm-Nature2013, Webb-Nature2018, Baker-Sci2022}, and materials sciences \cite{Stan-npj2018, Anthony-CM2020, Batra-NRM2021}, and has been widely used in various aspects of modern society, e.g., automatic driving systems, face recognition, fraud detection, expert recommendation system, speech enhancement, and natural language processing, etc. Especially, the deep learning techniques based on the artificial neural networks \cite{Hinton-Nature2015, GF-Book2016} have become the most popular and dominant machine learning approaches progressively, and their interactions with many-body physics have been intensively explored in recent years. On the one hand, some typical neural networks, such as multilayer perceptron \cite{GF-Book2016}, restricted boltzmann machine \cite{Smolen-Chapter1986}, autoencoder \cite{Kamp-BC1988}, convolutional neural network \cite{LeNet5-IE1998}, and autoregressive network \cite{Bengio-NIPS2000}, have been successfully applied to the study of quantum magnetization \cite{Troyer-Sci2017, CZ-PRB2018, HLX-PRB2018, Imada-PRX2021}, Fermi-Dirac statistics \cite{FermiNet-PRR2020, PauliNet-NC2020}, superconductivity \cite{NJ-npj2019, Yuki-PRB2021}, statistical averages \cite{WD-PRL2019} and phase transitions \cite{Melko-NP2017, Huber-NP2017, Wetzel-PRE2017} in physical systems. On the other hand, the ideas and techniques developed in physics are introduced into neural networks to improve the performance as well as interpretability \cite{Valiant-CACM1984, Ganguli-AR2020}. This approach might obtain a deeper insight of the neural networks and is sometimes referred to as the physics-inspired machine learning \cite{NML2021, YL-NRP2021}. A successful example is the introduction of tensor-network state into deep learning. It stems from quantum information and develops fastly in quantum many-body physics, and recently it has been used to realize the supervised learning \cite{Miles-NIPS2017, Cirac-IE2020, CS-PRB2021}, generative models \cite{ZP-PRX2018, CS-PRB2019, FV-arXiv2022}, and network reconstruction \cite{LD-NJP2019, GZF-PRR2020, Bojan-QMI2022}, etc. Though there are some limitations and difficulties in the current stage, it can still be expected that the interplay between deep learning and many-body physics will continue to flourish in the next decade.

Among the discussions about machine learning and many-body physics, the connection between deep neural networks and the renormalization group (RG) has been extensively studied in the literature recently \cite{Saeed-PNAS2013, Beny-arXiv2013, Lin-JSP2017, Schwab-arXiv2016, Janusz-NP2018, LW-PRL2018, CL-IE2020}. This relevance probably stems from the essential similarity between the underlying hierarchical structure of the inference process in supervised learning \cite{GF-Book2016} and the generated coarse-grained structure in the RG flow in physical systems \cite{CK-Book}, and can be seen more clearly in the context of tensor renormalization group, where the tensor-network structure and the RG-based techniques are combined together to study the many-body physics \cite{TRG-PRL2007, SRG-PRL2009, HOTRG-PRB2012, Kadanoff-RMP2014, Yannick-RMP2022}. In fact, it shows that not only the hierarchical structure but also the backpropagation method employed in the training process of neural networks resemble those of the tensor networks very much \cite{SRGAD-PRB2020}, and this actually lays the foundation of the increasing interplays between the two fields.

In this work, we propose to explicitly employ the hierarchical coarse-grained structure, as generated in the RG process in tensor networks similarly, in neural networks, and apply it to image classification and data distillation \cite{Hinton-arXiv2015, Wang-arXiv2018, Zhao-arXiv2020} for better interpretability in both physics and mathematics. To be specific, in the classification task, we construct a neural network called TaylorNet, which expresses the mapping from the input data to the output label in terms of the Taylor series approximately without using any nonlinear activation functions. The network is simple and can be expressed as a polynomial manifestly in mathematics. This is very different from the ordinary neural networks whose explicit expressions are difficult to obtain, thus unveiling part of the mysteries of neural networks and providing clear direction for further improvement. In the second part, we design a multi-level distillation process which imitates the coarse-graining (CG) operations in the RG flow and displays the underlying hierarchical structure of the inference process explicitly. It shows that the data distilled from lower-level abstraction contains much more details than those distilled from higher-level abstraction, and the final data distilled from the highest-level layer can be used as good references for image classification directly. The results obtained in both tasks are rather satisfying, as demonstrated in the MNIST \cite{MNIST} and CIFAR-10 \cite{CIFAR} datasets.

The rest of the paper is organized as follows. In Sec.~\ref{sec:CG}, we briefly review the coarse-grained structure generated in the RG process in the context of the matrix product operator. In Sec.~\ref{sec:TaylorNet} and Sec.~\ref{sec:DataDist}, we introduce the TaylorNet and the new setup for data distillation, respectively, and demonstrate their validity in MNIST and CIFAR-10 datasets. In Sec.~\ref{sec:Summary}, we summarize our work and discuss the possible improvement as well as promising extensions briefly.

\section{Coarse-grained structure generated in the RG process}
\label{sec:CG}
The RG is one of the most profound tools conceptually of theoretical physics \cite{GL-PR1954, Kadanoff-Phy1966, Wilson-PRB1971, Wilson-RMP1975, Kadanoff-PRL1975}, and its impact spans from high-energy to statistical and condensed matter physics \cite{GL-PR1954, Swend-PRL1979, White-PRL1992, Wet-PLB1993}. Essentially, the RG is a conceptual framework comprising various techniques, such as the original block spin approach \cite{Kadanoff-Phy1966, Kadanoff-RMP2014}, functional RG \cite{Wet-PLB1993}, Monte Carlo RG \cite{Swend-PRL1979}, density matrix RG \cite{White-PRL1992}, and tensor network RG \cite{TRG-PRL2007, SRG-PRL2009, HOTRG-PRB2012, Yannick-RMP2022}, and so on. Though these schemes differ substantially in details, they share a same essential feature, namely the RG process aims to identify the relevant degrees of freedom (DOFs), integrate out the irrelevant ones iteratively, and eventually arrive at a low-energy effective theory. The extraction of the relevant information is realized by a set of RG transformations, which map the DOFs in a lower scale to those in its neighboring higher scale. A hierarchical coarse-grained structure is essentially generated during this kind of scale transformation. It can be seen more clearly in the real-space RG schemes \cite{Kadanoff-Phy1966, Kadanoff-RMP2014, White-PRL1992, TRG-PRL2007, SRG-PRL2009}, as exemplified in the following.

\begin{figure}[htbp]
	\includegraphics[width=0.3\textwidth]{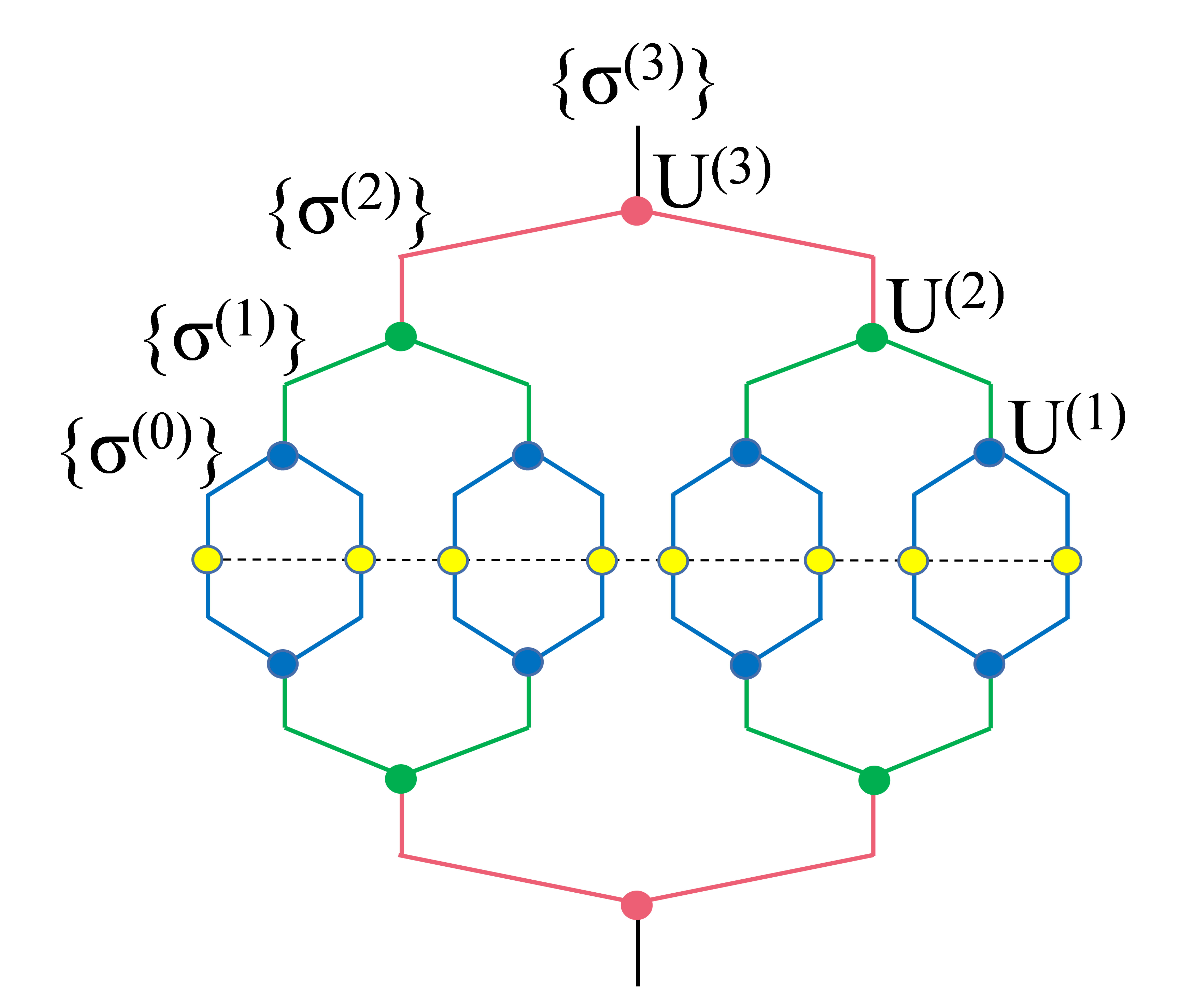}
	\caption{Hierarchical coarse-grained structure generated in the RG process for a matrix product operator with open boundary condition. The hollow circles connected by a dashed line and colored by yellow denote the lattice sites where the operator is defined. As described in the main text, the local RG transformations are represented by the rank-3 tensors $U$s denoted by solid dots, and the DOFs reside on the links between the dots and are denoted as $\{\sigma\}$. For the sake of clarity, the various scales are distinguished by different colors.}
	\label{fig:RG}
\end{figure}

To show the coarse-grained structure mentioned above clearly, let's consider the real-space RG transformations of a matrix product operator \cite{FV-PRL2004, FV-NJP2010} defined on a one-dimensional lattice, which may represent a many-body Hamiltonian of a quantum system or a transfer matrix of a classical statistical model. In this context, through a series of scale transformations, the RG process aims to find a finite-dimensional representation of the operator, which can preserve the low-energy part of the Hamiltonian or the dominant-eigenvalue part of the transfer matrix approximately. For simplicity, let's assume the MPO is symmetric, and consider the simple case where a binary mapping is performed in each scale transformation. Then the RG process in such a system with length $L=8$ and open boundary condition can be illustrated in Fig.~\ref{fig:RG}. At the beginning, the operator is expressed in terms of the variables $\{\sigma^{(0)}\}$ sitting on the blue lines. The 1st scale transformation is composed of four isometries denoted as $U^{(1)}$s, each of which maps $\{\sigma^{(0)}\}$ sitting on two neighboring blue lines to the variables $\{\sigma^{(1)}\}$ sitting on the corresponding green lines. Similarly, the 2nd scale transformation is composed of two isometries $U^{(2)}$s, and each $U^{(2)}$ maps $\{\sigma^{(1)}\}$ to variables $\{\sigma^{(2)}\}$ sitting on the red lines. $U^{(3)}$ constitutes the last scale transformation, and maps $\{\sigma^{(2)}\}$ to variables $\{\sigma^{(3)}\}$ sitting on the black lines. Eventually, the operator is represented in terms of $\{\sigma^{(3)}\}$, and this completes the full RG process. 

In the RG transformations described above, by isometry we mean there are fewer DOFs after the mapping, and the generated variables and DOFs are usually termed as coarse-grained. As clearly sketched in Fig.~\ref{fig:RG}, the whole RG process generates a hierarchical coarse-grained structure with three levels. For a given level, the RG transformations identify the relevant DOFs from lower-level DOFs, and output the identified DOFs to a higher-level transformation for further extraction. This CG operation is an essential gradient of the RG process, and has been heavily employed in the original block spin numerical RG calculations \cite{Chui-PRB1979, CXY-PRB1987, Kovarik-PRB1990} and the more recent tensor network RG proposals \cite{HOTRG-PRB2012, YLP-PRE2016, BBChen-PRX2018}. 

Without discussing the RG flow in the parameter space and the corresponding fixed-point properties, in this work, we just focus on the hierarchical coarse-grained structure described above and emphasize its similarity to the inference process in supervised learning tasks like image classification. Deep learning falls in the category of representation learning, whose central task is to extract high-level abstract features relevant to the final target from the raw data possessing many irrelevant details and variations \cite{GF-Book2016}, and it solves this problem by constructing higher-level representations out of simpler lower-level representations. More specifically, a representation in a given level is characterized by the output of the previous lower-level neural network layer, and is regarded as the input of a new hidden layer to construct the more abstract higher-level representation. The desired representation is eventually obtained by multistep abstraction, each step of which is realized by a hidden layer and extracts increasingly abstract features from the original input data. This multistep abstraction process is very similar to the RG process discussed before and illustrated in Fig.~\ref{fig:RG}, and also shows an underlying hierarchical coarse-grained structure. This similarity is more evident for the convolutional neural network, where the local structure is emphasized by convolution operations \cite{LeNet5-IE1998}. 

In the following sections, we introduce this hierarchical coarse-grained structure into neural networks manifestly, which makes the conceptual similarity described here more explicit, and the resulting networks are much more easier to understand in both physics and mathematics.

\section{TaylorNet}
\label{sec:TaylorNet}
The quintessential example of a deep learning model is the deep feedforward neural network, and sometimes is referred to as multilayer perceptron model \cite{GF-Book2016}. It represents a mapping $\mathcal{F}$ from the input data to the output result, and generally can be expressed as a composite function of many linear $(\mathcal{L})$ and nonlinear $(\mathcal{N})$ transformations ordered alternately. For example, a feedforward neural network with $n$ layers can be represented as
\begin{equation}
	\mathcal{F} = \mathcal{N}_n\mathcal{L}_n\cdots\mathcal{N}_2\mathcal{L}_2\mathcal{N}_1\mathcal{L}_1 	\label{eq:FNL}
\end{equation} 
where the linear mappings $\mathcal{L}$s contain many variational parameters that need to be determined, while the nonlinear mappings $\mathcal{N}$s contain almost no free parameters and are realized by some known operations called activations, such as rectified linear unit, logistic sigmoid, maxpoolings, and so on. The nonlinear mappings $\mathcal{N}$s are indispensable to approximate a nonlinear $\mathcal{F}$ \cite{GF-Book2016}. Apparently, Eq.~(\ref{eq:FNL}) seems oversimplified, but fortunately, when $\mathcal{F}$ is Borel measurable, the validity is guaranteed by the universal approximation theorem \cite{Hornik-NN1989, Cybenko-MCSS1989}, as long as the neural network is sufficiently wide and at least one $\mathcal{N}$ is squashing in some sense. Therefore, Eq.~(\ref{eq:FNL}) provides a quite general belief to approximate an actual mapping in practical applications of neural networks. However, for a given mapping $\mathcal{F}$, generally, there is no clue on either how wide the neural network is or how we can obtain the desired $L$s. In order to determine the parameters effectively, much effort has been devoted to designing special structures, and this greatly boosted the development of deep neural networks. Successful structures include the convolution operation \cite{LeNet5-IE1998}, shortcut connection in residual network \cite{HKM-CVPR2017, HG-CVPR2017}, attention structure in the transformer model \cite{Ashish-NIPS2017}, etc. Nevertheless, the specific design of structures relies mainly on empirical experience, and there is no theoretical guidance generally, and this is why deep learning is usually regarded as a magic black box.

To reduce the mystery in the structure design, in this work, we propose to use another universal expansion, i.e., the multi-variable Taylor formula valid for an arbitrary analytic function $f$. Expanded at a certain point, the expression can be written as
\begin{align}
	f(X) =& f_0 + \sum_{i=1}^{N}a^{(1)}_ix_i + \sum_{i,j=1}^{N}a^{(2)}_{ij}x_ix_j + \nonumber \\
	&\sum_{i,j,k=1}^{N}a^{(3)}_{ijk}x_ix_jx_k + \cdots, \label{eq:Taylor}
\end{align}
where $X$ is the input vector with $N$ elements denoted as $\{x_1,x_2,...,x_N\}$, $a^{(n)}$ is the coefficient related to the corresponding $n$-th order derivative, and $f_0$ is a collected constant. Eq.~(\ref{eq:Taylor}) is also universal, since the nonanalytic functions encountered in our daily life are always expected to have a finite number of singular points and thus can be well approximated by Eq.~(\ref{eq:Taylor}) arguably. Hereafter, just for convenience, we simply refer to $a$ and $x_ix_jx_k...$ as the Taylor coefficient and Taylor term, respectively. 

The validity of Eq.~(\ref{eq:Taylor}) can be directly verified by an experiment on the classification of MNIST dataset. In the experiment, we regard each image as a vector $X$, and assume $f^{(\alpha)}(X^{(i)})$ can be expanded as Eq.~(\ref{eq:Taylor}), where $f^{(\alpha)}(X^{(i)})$ denotes the probability of the $i$-th image belongs to the $\alpha$-th category. The whole neural network has only a linear layer that holds the coefficients, and the output can be expressed as
\begin{align}
	f^{(\alpha)}(X^{(i)}) = \sum_{j=1}^p W_{\alpha,j}\tilde{X}^{(i)}_j.  \label{eq:NaTay}
\end{align}
In the above equation, $\tilde{X}^{(i)}$ is a vector containing $p$ individual Taylor terms corresponding to $X^{(i)}$ which are retained in Eq.~(\ref{eq:Taylor}), and $W$ is a 10$\times p$ weight matrix with element $W_{\alpha,j}$ the Taylor coefficient corresponding to $\tilde{X}^{(i)}_j$. To make the calculation feasible, we resize the original 28$\times$28 images into 7$\times$7 through the well-established bilinear interpolation technique \cite{William-Book2007}, do expansion up to the fourth order, and collect all possible terms in $\tilde{X}$.

\begin{figure}[htbp]
	\includegraphics[width=0.4\textwidth]{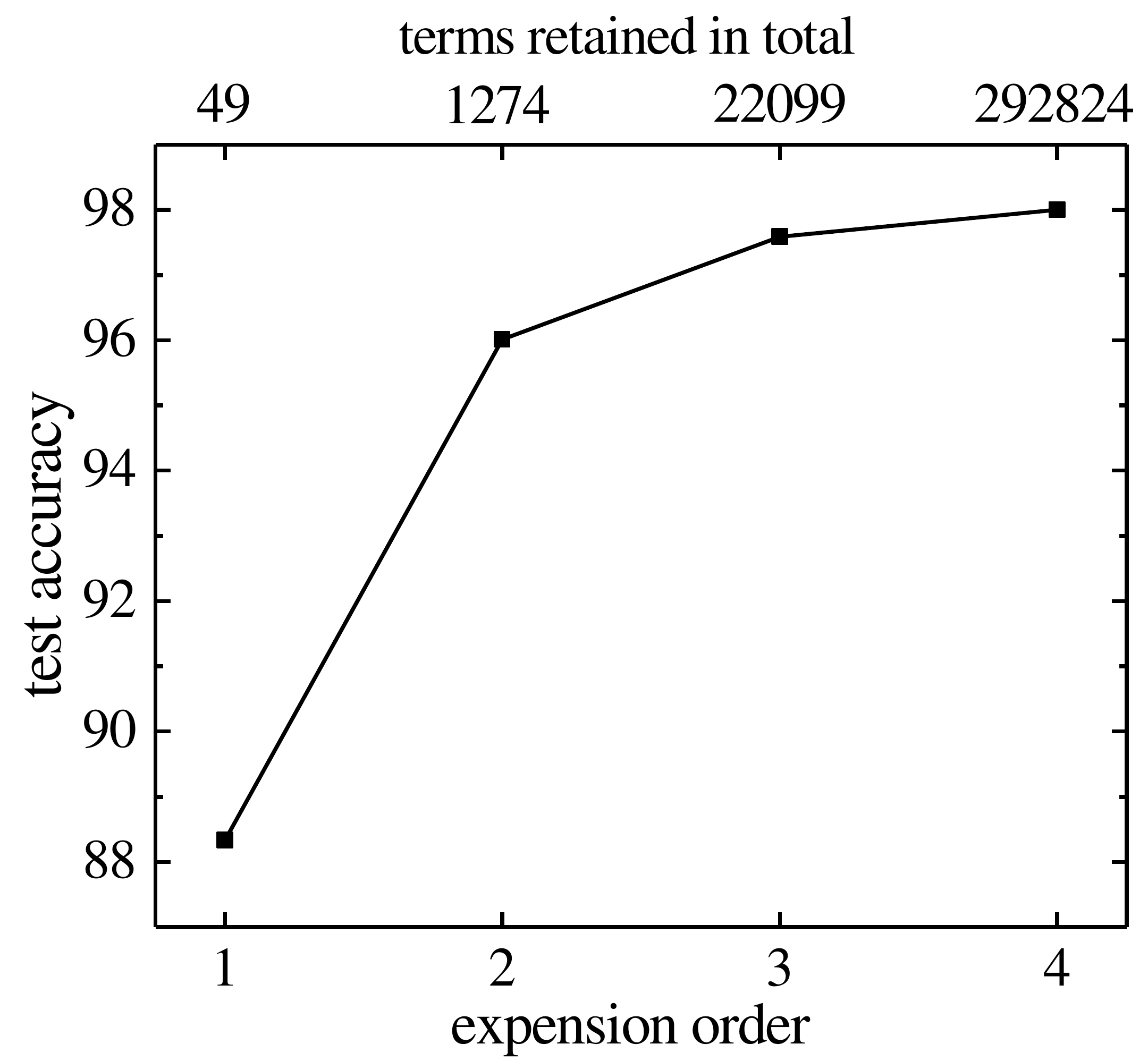}
	\caption{Test accuracy of the experiment on direct Taylor expansion, as expressed in Eq.~(\ref{eq:NaTay}). The data is obtained on the MNIST dataset with resized $7\times7$ images.}
	\label{fig:TaylorAcc}
\end{figure}

\begin{figure}[htbp]
	\centering
	\subfigure[]{
		\begin{minipage}[t]{0.48\linewidth}
			\centering
			\includegraphics[width=\linewidth]{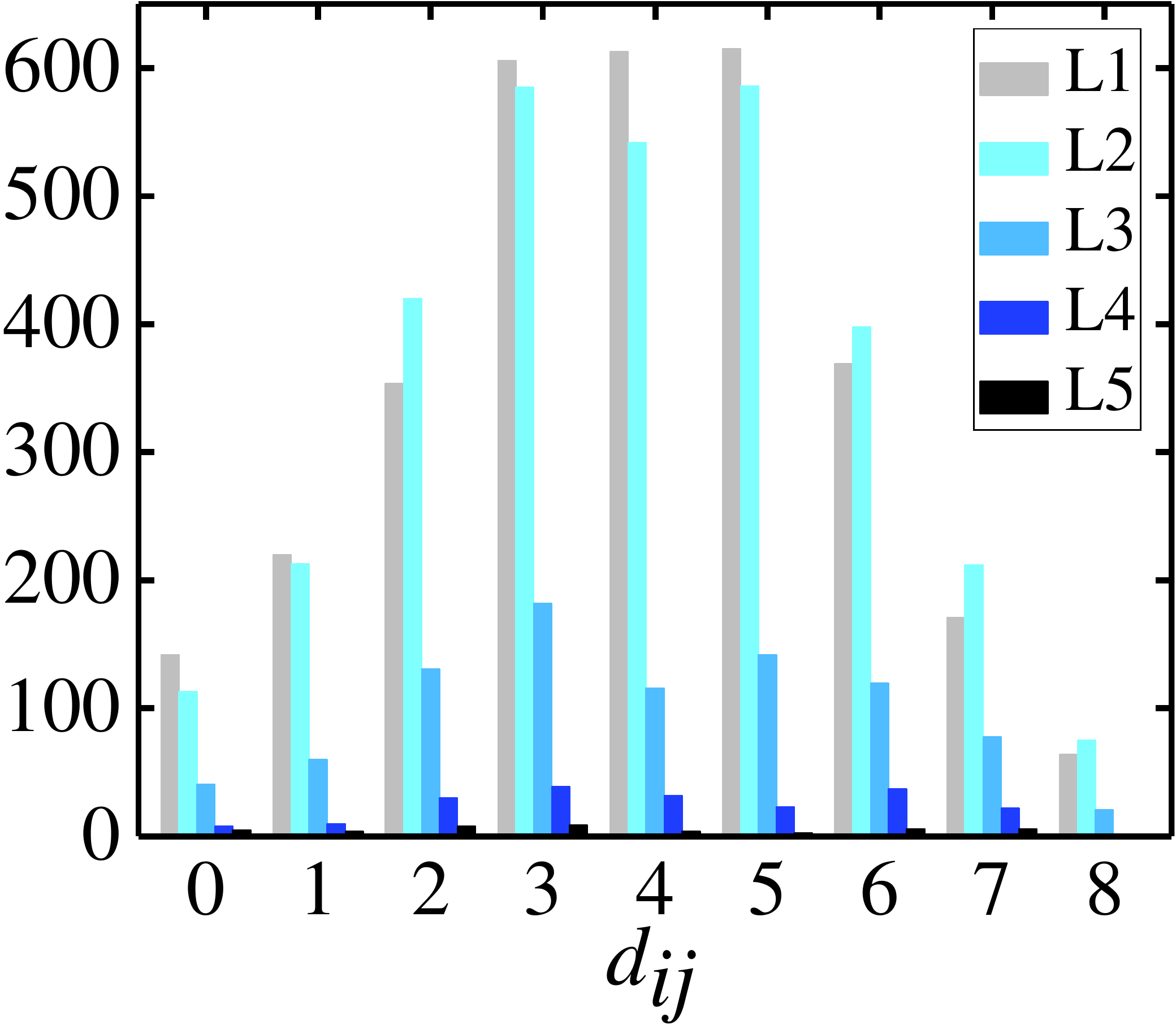}
		\end{minipage}%
	}%
	\subfigure[]{
		\begin{minipage}[t]{0.49\linewidth}
			\centering
			\includegraphics[width=\linewidth]{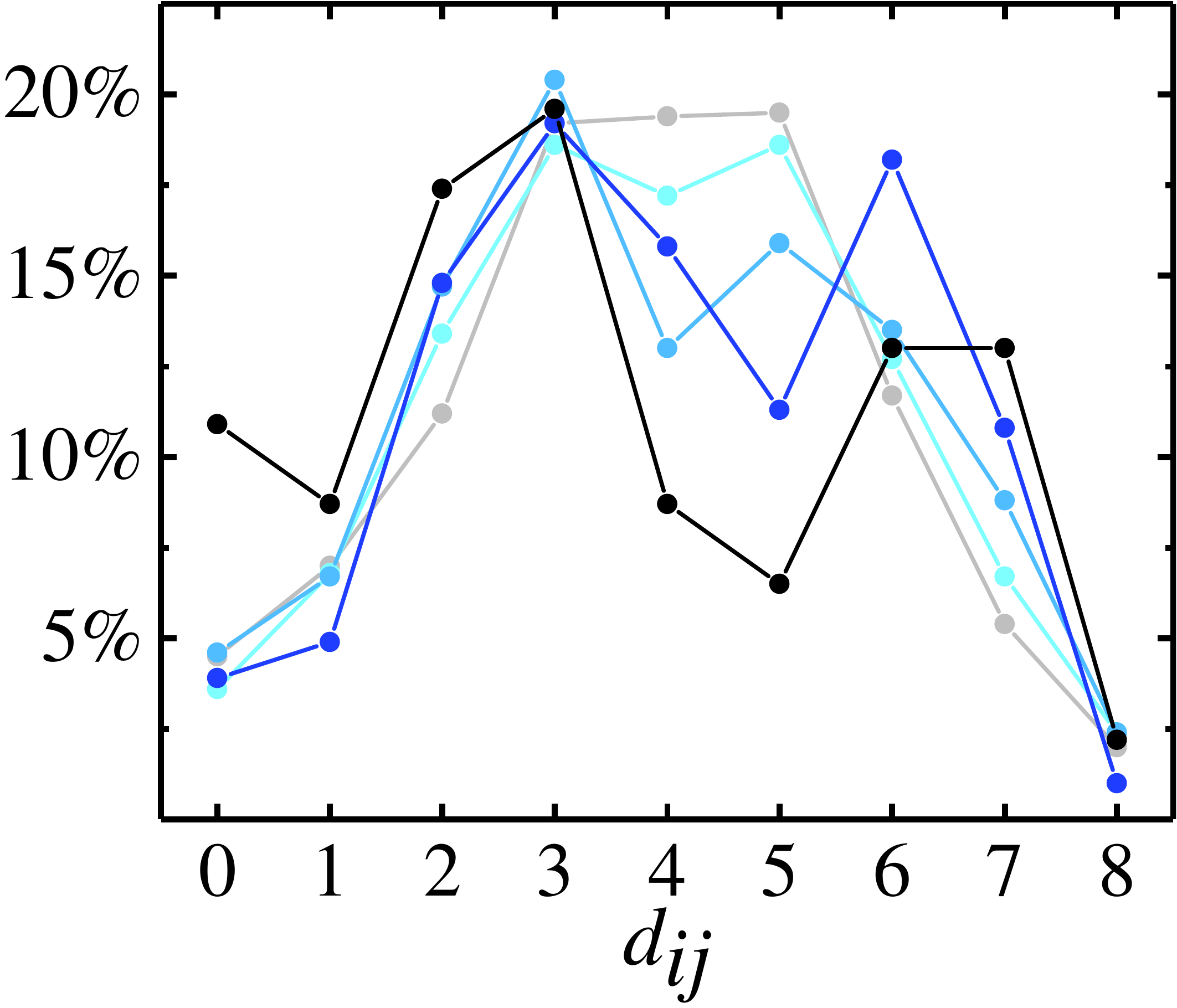}
		\end{minipage}
	}%
	\caption{Distribution of the obtained weight corresponding to the quadratic terms $x_ix_j$, as a function of distance $d_{ij}$ between the two pixels in images of the 7$\times$7 MNIST dataset. Weights are ordered by value in ascending order, and equally divided into five groups that are referred to as level-1 (L1) to level-5 (L5), respectively. (a) Weight distribution at each distance $d_{ij}$. (b) Weight ratio distribution for each level as a function of $d_{ij}$. The details of the distribution can be found in Fig.~\ref{fig:DistribDetail} in App.~\ref{app:Weight}.}
	\label{fig:Distrib}
\end{figure}

The result is shown in Fig.~\ref{fig:TaylorAcc}. It is clear that the test accuracy can be systematically improved as the expansion order $n$ is increased. When $n=4$, the number of total terms retained is about 293 thousand, and the obtained accuracy is about 98\%. As a comparison, on the same 7$\times$7 MNIST dataset, a residual network with 1.3 million parameters can achieve an accuracy of about 99\%. The performance can be further improved by some detailed analysis. Especially, it shows that, though there are about totally 293 thousand terms retained in the expansion, the contribution of a great number of terms is very small. For example, the distribution of weight corresponding to the quadratic terms is displayed in Fig.~\ref{fig:Distrib}. For a given term $x_ix_j$, Fig.~\ref{fig:Distrib}(a) tells the dominant weight always consumes the least portion no matter how far $x_i$ and $x_j$ are separated, and Fig.~\ref{fig:Distrib}(b) tells that all the weights have a preferred distribution as a function of the distance between $x_i$ and $x_j$. This reminds us that there is much redundancy in $W$, and thus the number of parameters, $N_0$, can be greatly reduced by discarding the small weights. In fact, experiments show that the accuracy remains unchanged when $N_0$ is reduced by half, and drops only by 0.83 percent when $N_0$ is reduced to 30\%. Even if $N_0$ is reduced to 10\%, we can still obtain an accuracy of about 85\%. This actually reflects the spirit of Taylor expansion, since it means the accuracy can be systematically improved by adding more subtle terms.

\begin{figure}[htbp]
	\centering
	\includegraphics[width=0.45\textwidth]{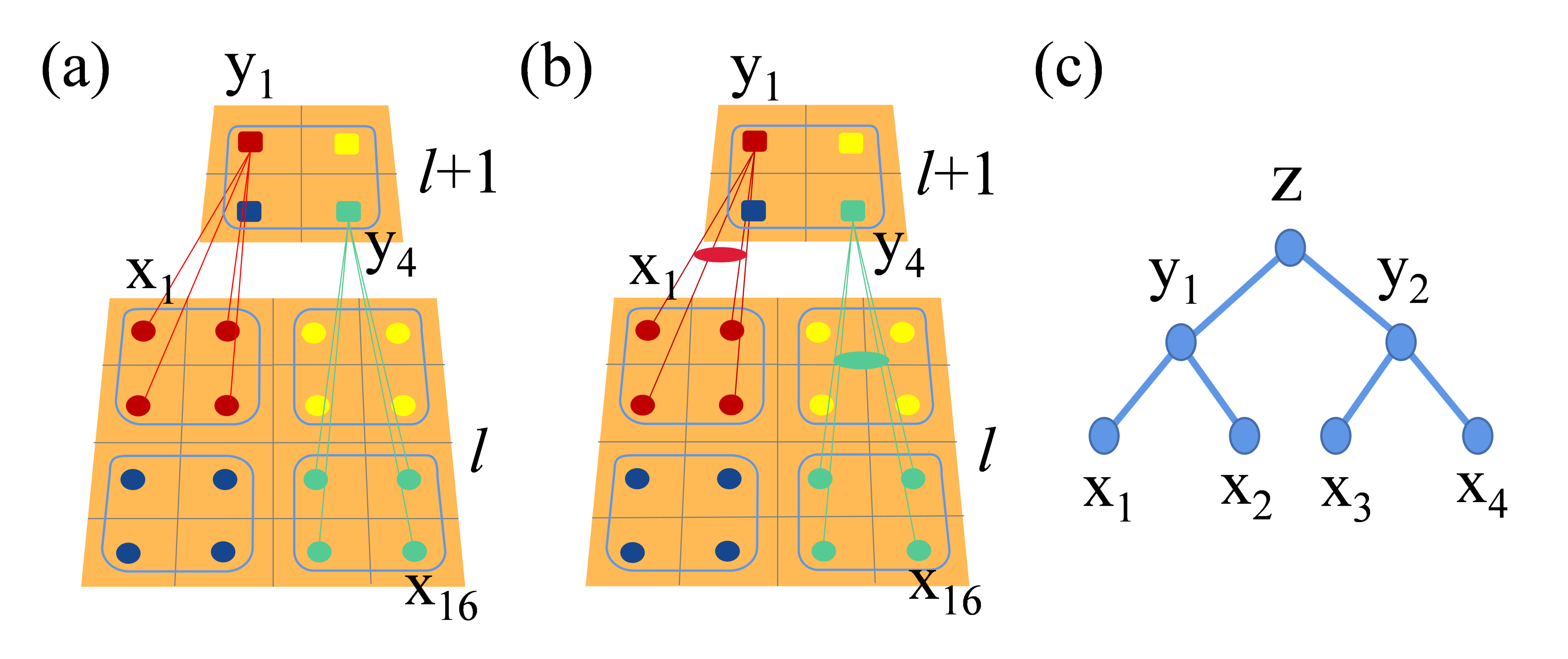}
	\caption{Illustration of the CG operation in TaylorNet. (a) Convolution operation. (b) CG operation. Clearly, the product of $x_1$ and $x_{16}$ will emerge in the next CG after the next CG operation. (c) Two successive CG operations on four variables divided into two local clusters. Very complicated terms of $x$ emerge in the expression of $z$, as shown in Fig.~\ref{fig:W} in App.~\ref{app:Yexp}.}
	\label{fig:CGcore}
\end{figure}

To extend Eq.~(\ref{eq:Taylor}) to large scale computer tasks, and to make the above procedure more practical and efficient, we propose the TaylorNet, which realizes Eq.~(\ref{eq:Taylor}) in a multistep manner by utilizing the hierarchical coarse-grained structure described in Sec.~\ref{sec:CG}. Based on the assumption that the dominant parts in the Taylor series mainly correspond to the product of $x$s in local clusters, as has been partially evidenced in Fig.~\ref{fig:Distrib}, we introduce a series of intermediate variables $x_{\alpha}$ with the new index $\alpha$ indicating the hierarchical levels. The variables at a higher level are to be expressed in terms of Taylor series with respect to the variables at the neighboring lower level, and constitute the Taylor expansions of the variables at the neighboring higher level. To be specific, if expanded to the second order, a local cluster indexed as $\alpha$ with $m$ $n$-th level variables denoted as $\{x^{(n)}_1, x^{(n)}_2, ..., x^{(n)}_m\}$ is mapped to a variable $x^{(n+1)}_\alpha$ at the ($n$+1)-th level, i.e.,
\begin{align}
	x^{(n+1)}_\alpha = & c^{(n+1,0)} + \sum^{m}_{i=1}c^{(n+1,1)}_ix^{(n)}_i + \cdots \nonumber \\
	                   &\sum_{i,j=1}^{m}c^{(n+1,2)}_{ij}x^{(n)}_ix^{(n)}_j 	\label{eq:LocalMap}
\end{align}
where $c^{(n,\sigma)}$ denotes the coefficients introduced in the $\sigma$-th order terms in the expansion of variables at the $n$-th level, and $x^{(0)}_i$ is defined as the original input data $x_i$. Hereafter, Eq.~(\ref{eq:LocalMap}) is referred to as a CG operation, and it is illustrated in Fig.~\ref{fig:CGcore}, in which the usual convolution operation is also illustrated for comparison. Suppose we are considering the simplest case, i.e., the sizes of the local cluster is 2$\times$2, and there is no overlap between the clusters. In the language of neural networks, this means the size of both the kernel and the stride is 2$\times$2. In Fig.~\ref{fig:CGcore}(a) and (b), the variables at two neighboring levels are denoted as dots and squares, respectively, and the variables associated with a single local mapping are indicated by the same color. In a convolution operation, a square is a linear combination of four dots, which corresponds to the linear terms in Eq.~(\ref{eq:LocalMap}). While in the CG operation, a square is a nonlinear combination of the same four dots, which corresponds to Eq.~(\ref{eq:LocalMap}) exactly. To indicate the nonlinear feature, an oval plate is added to distinguish from the convolution, as shown in Fig.~\ref{fig:CGcore}(b).

The local introduction of the nonlinear mapping has a great advantage over the initial proposals of Taylor expansion, as expressed in Eq.~(\ref{eq:Taylor}) and Eq.~(\ref{eq:NaTay}). On the one hand, the locality puts a strong constraint on the distance among the variables showing up in the Taylor terms retained in the expansion. This greatly reduces the number of parameters that need to be determined, and also removes some unnecessary redundency, since the contribution from the terms involving variables faraway separated is expected to be small, as partially evidenced in Fig.~\ref{fig:Distrib}(b). On the other hand, the higher-order terms, as well as the terms involving variables belonging to different local clusters, can emerge naturally in the next several CG operations, which can be seen from Fig.~\ref{fig:CGcore}(c) explicitly. For example, more complex terms like $x_i^3$, $x_i^4$, $x_1x_3$, $x_1x_2x_3$, $x_1x_2x_3x_4$ show up in the expression of $z$, though $y_1$ and $y_2$ are only expanded to the second order locally. The full expression of $z$ can be found in App.~\ref{app:Yexp}. Thus by introducing the nonlinear terms locally, we can generate long-range and very complicated terms in the actual expansions of the variables at the highest level, and there is no need to invoke any magic activations at all.

\begin{figure}[htbp]
	\centering
	\includegraphics[width=0.5\textwidth]{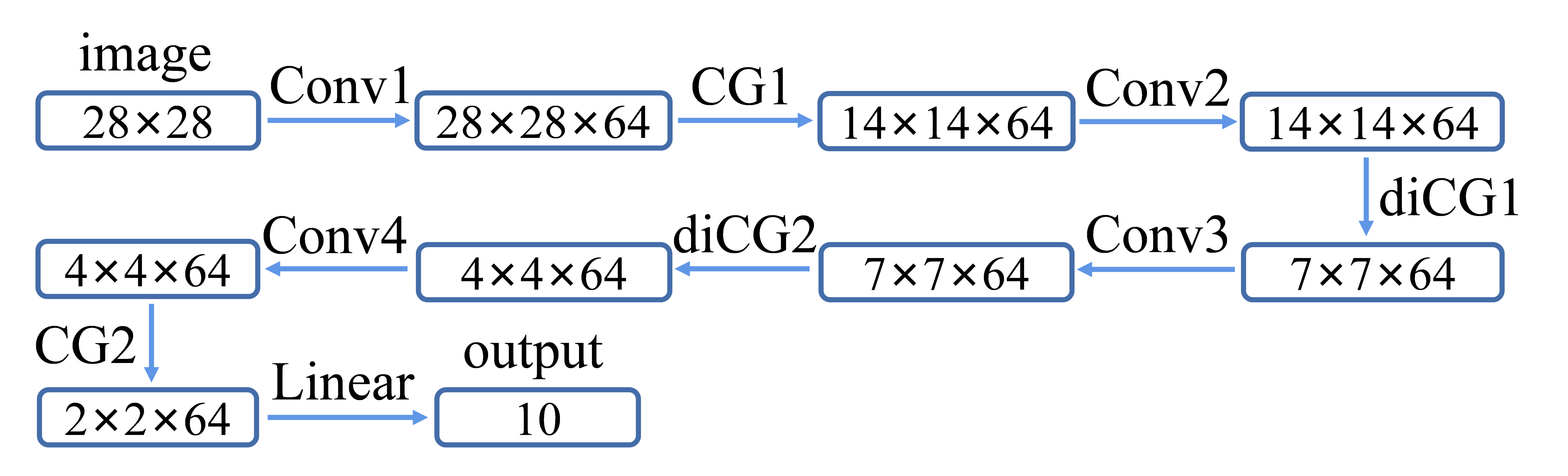}
	\caption{Sketch of the TaylorNet used in the classification task on MNIST dataset. Hereafter, the numbers in the box denote the representation form of the data, e.g., $28\times 28\times 64$ denotes 64 feature maps with size $28\times 28$, and the operations sit on the arrows correspond to different neural network layers, e.g., Conv($l_1$,$l_2$,c,$s_1$,$s_2$,p) means convolutional layer with kernel size $l_1\times l_2$, channels $c$, stride size $s_1\times s_2$, and padding number $p$ (default 0), similar for CG operation and dilated CG operation as discussed in the main text and Fig.~\ref{fig:DiCG}. Here, all the four convolutional layers have structure Conv(3,3,64,1,1,1), both CG layers have structure CG(2,2,64,2,2), diCG1 and diCG2 have structures CG(2,2,64,1,1,7) and CG(2,2,64,1,1,3), respectively. The action of a multi-channel convolution operation is illustrated in Fig.~\ref{fig:ConvOP} in App.~\ref{app:Conv}, and more details can also be found in Ref.~\cite{GF-Book2016}.}
	\label{fig:NetStr}
\end{figure}

In our experiment on MNIST, we use a TaylorNet with four CG layers, each of which maps a 2$\times$2 cluster to a single variable according to second-order Taylor expansion, and then use a linear layer that maps the resulting 2$\times$2 variables to a vector with 10 elements representing the probabilities. The detailed TaylorNet structure is shown in Fig.~\ref{fig:NetStr}. The obtained accuracy is about 99.2\%, which is quite satisfying. On the resized 7$\times$7 MNIST dataset, we can obtain an accuracy of about 97.9\% with about 248 thousand parameters in total, which is much less than the parameters in both the original Taylor expansions (1.46 million) and the residual network (1.3 million) described before. As to the CIFAR-10 dataset, we can obtain an accuracy of about 71.7\% with only 1.2 million parameters, and this is also more efficient than the recent MLP-Mixer proposal \cite{Mixer-arXiv2021} without pre-training process, which combines the information from the inter-cluster and intra-cluster variables in a similar way. The detailed TaylorNet structure for CIFAR-10 dataset can be found in in Fig.~\ref{fig:NetStrCifar} in App.~\ref{app:NetStr}. Furthermore, it shows that if we replace the CG operations with the convolution operations in the whole neural network, the accuracy will drop by about 7.3\% and 30\% immediately as expected, on MNIST and CIFAR-10 datasets, respectively. This clearly demonstrates the power of CG operations in the representations of nonlinear mappings. 

Similar to the convolution operation, the above CG operation can be performed in slightly different manners. Firstly, the size of the local clusters can be different, and there can be overlaps between different clusters. Moreover, the translation and/or scale invariance of the CG kernels can be employed, that is, the Taylor coefficients in Eq.~(\ref{eq:LocalMap}) for CG operations performed at different clusters and/or in different scales can be assumed identical. Secondly, the expansion order in Eq.~(\ref{eq:LocalMap}) can be larger than 2, and this depends on the strength of the nonlinearity in the expansion of the actual $\mathcal{F}$.

\begin{figure}[htbp]
	\centering
	\includegraphics[width=0.25\textwidth]{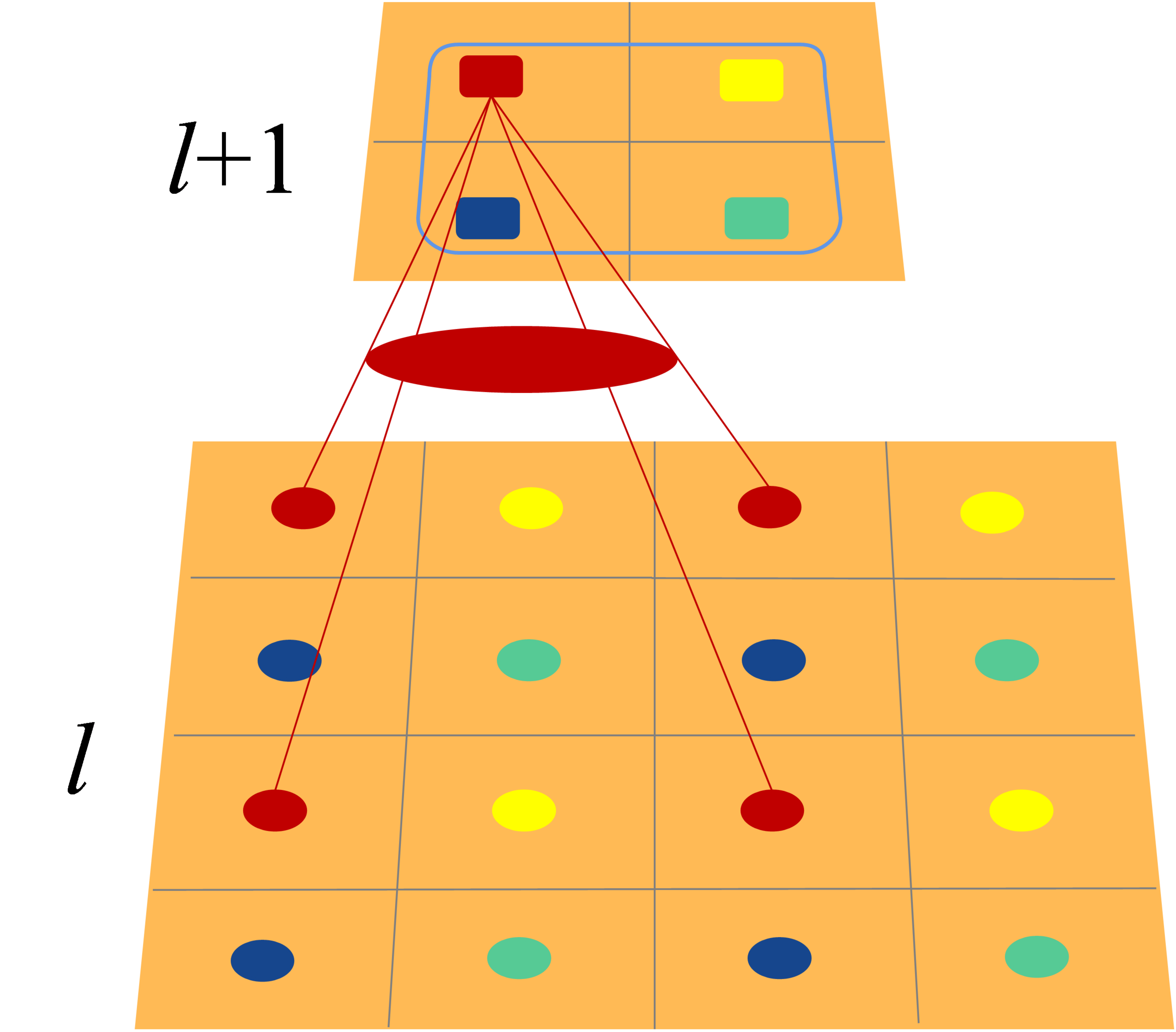}
	\caption{Sketch of a dilated CG operation, used in Fig.~\ref{fig:NetStr}. The structure shown in this figure is denoted as CG(2,2,$n$,1,1,2) which means the kernel size is $2\times 2$, number of channels is $n$, stride size is $1\times 1$, and the dilation is 2 in both directions.}
	\label{fig:DiCG}
\end{figure}

It is worth mentioning that, in Fig.~\ref{fig:NetStr}, we have employed two simple ways to further enlarge the effective receptive fields \cite{Luo-arXiv2017} of the CG operations, without changing the size of the local clusters manifestly. One is the introduction of the dilated CG operation, as shown in Fig.~\ref{fig:DiCG}. In the dilated CG, the variables need to be coarse grained scatter separately in different clusters instead of aggregating locally, which is very similar to the structure of the dilated convolution \cite{Garrison-arXiv2017}. The other is performing convolution before the CG operation. This is easy to undertand, since the convolution turns each dot in Fig.~\ref{fig:CGcore}(b) as the linear combination of several dots nearby before the CG operation, and thus each square is effectively expressed in terms of more dots actually. These two techniques might be advantageous in some situations where the inter-cluster product is more important in Eq.~(\ref{eq:NaTay}), and are also considered in Ref.~\cite{Mixer-arXiv2021} similarly.

\section{Data distillation}
\label{sec:DataDist}
The concept of knowledge distillation was originally proposed by Hinton \textit{et al.} \cite{Hinton-arXiv2015}, and it aims to train a simpler neural network called student, which is expected to have the same performance approximately to a complex model referred to as teacher. Later, data distillation is proposed to train a smaller dataset from a larger dataset, and expect that the obtained distilled dataset can be used to efficiently train a neural network that has a similar performance to that of a neural network trained from the original larger dataset \cite{Wang-arXiv2018, Zhao-arXiv2020}. Though the process of distillation is somewhat complicated, the idea is very simple and reasonable, namely the neural-network-based deep learning is believed to be able to extract some essential features from the original dataset.

In order to make the above idea clearer, and display the inference or abstraction process more explicitly, we propose a new setup of data distillation. Utilizing the hierarchical coarse-grained structure, the new proposal aims to extract the essential features through a multistep process, in which the abstraction is performed progressively from lower levels to higher levels. This is actually the essential spirit of deep learning \cite{GF-Book2016}, as discussed in Sec.~\ref{sec:CG}. It shows that the distilled dataset can be indeed used as references to perform classification task directly. 

\begin{figure}[htbp]
	\includegraphics[width=0.4\textwidth]{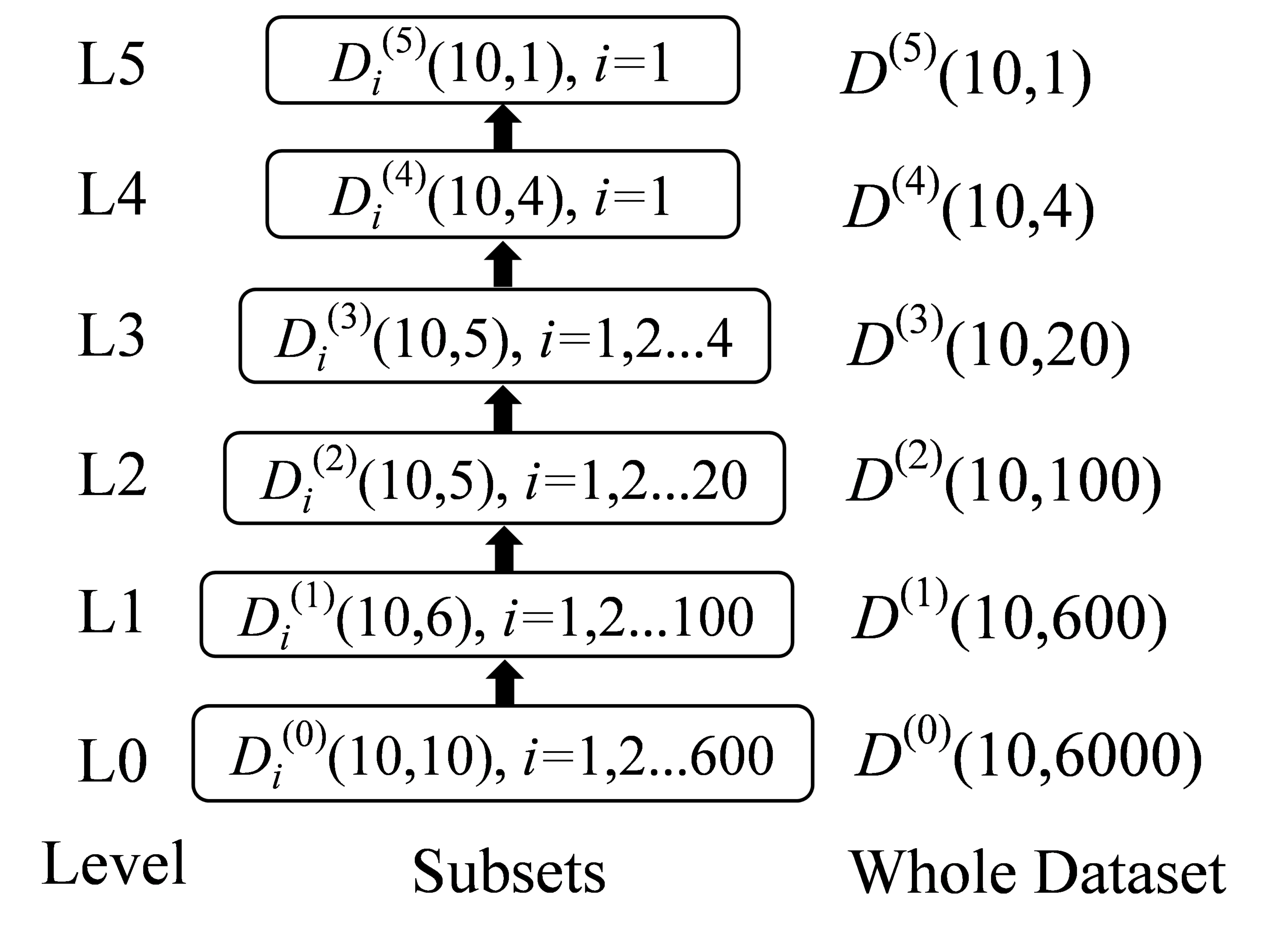}
	\caption{The data distillation strategy described in the main text, designed for MNIST dataset. The whole distillation process has a five-level structure. The correspondence in the five distillation procedures can be represented as 10-to-1, 6-to-1, 5-to-1, 5-to-1, and 4-to-1 mappings, respectively.}
	\label{fig:DDstrategy}
\end{figure}

For concreteness, in the following we describe the distillation strategy applied to the MNIST dataset. The original dataset is composed of ten classes, each of which contains 6000 images, and is denoted as $D^{(0)}(10,6000)$ hereafter. Firstly we divide each class into 600 groups equally, and select one group from each class to constitute a subset which contains 10 images for each class. Thus totally we obtain 600 subsets, and for simplicity, the $i$-th subset is denoted as $D^{(0)}_i(10,10)$ and has 100 images in total. Then perform the usual distillation process on each subset $D^{(0)}_i$ by neural networks, as will be described later, and distill 10 images corresponding to the 10 classes out of each subset. This completes the first level distillation procedure, from which 600 images for each class are distilled, and we denote the distilled dataset as $D^{(1)}(10,600)$ as a whole. Similarly, we further divide $D^{(1)}(10,600)$ into 100 subsets $D^{(1)}_i(10,6)$ on each of which the distillation process is performed and 10 distilled images are obtained, and then we obtain the distilled dataset $D^{(2)}(10,100)$ which contains 100 images for each class at the second level. Repeat this divide-and-conquer strategy, we can obtain dataset $D^{(3)}(10,20)$, $D^{(4)}(10,4)$, and finally reach the highest distilled dataset $D^{(5)}(10,1)$, which contains only a single distilled image for each class and can be regarded as the typical representatives hosting the essential features of that class. The whole process is illustrated in Fig.~\ref{fig:DDstrategy}.

\begin{figure}[htbp]
	\includegraphics[width=0.5\textwidth]{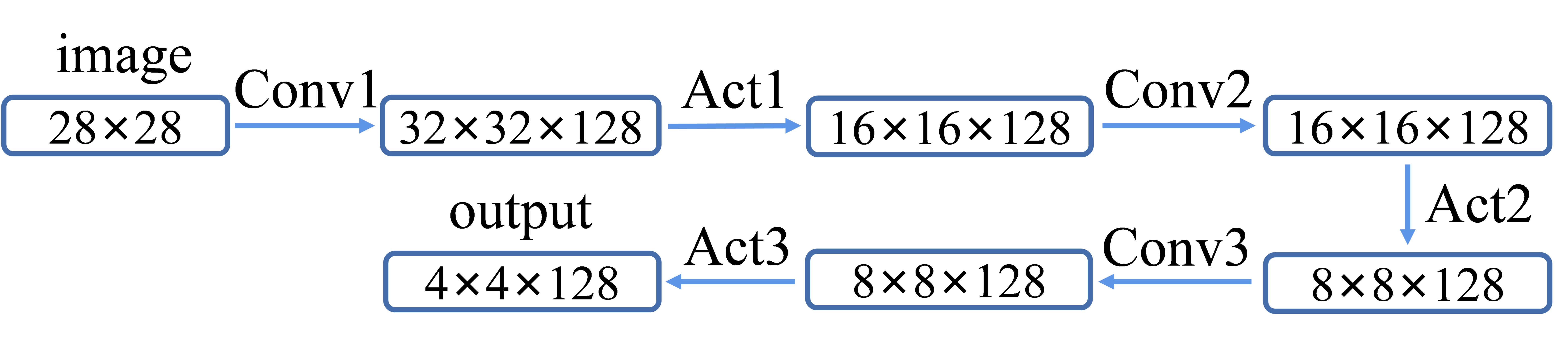}
	\caption{The neural network structure used in the distillation procedure on each subset of MNIST. Conv1 has structure Conv(3,3,128,1,1,3), both Conv2 and Conv3 have structure Conv(3,3,128,1,1,1). Act1, Act2, Act3 are three activation layers, each of which is composed of instance normalization, relu, and avgpooling. The avgpooling is performed with kernel size $2\times 2$ and stride $2\times 2$. Each image is mapped to an embedded space, and the result is a vector with length $16$ and channels $128$.}
	\label{fig:DDnet}
\end{figure}

In this work, to perform the distillation procedure on each subsets $D^{(\alpha)}_i(10,m)$, we employ the distribution matching method and a similar neural network architecture proposed in Ref.~\cite{Zhao-arXiv2020}, as is illustrated in Fig.~\ref{fig:DDnet} in detail. The goal of the procedure is to determine $n$ (distilled) images, denoted as $Y$s, which minimize a lost function defined in the following
\begin{align}
	L &= \sum_{\alpha=1}^n \lbra{\lambda d_{\alpha, \alpha} - \sum_{\beta\neq\alpha}^nd_{\alpha, \beta}} \nonumber \\
	d_{\alpha,\beta} &\equiv \sum_{i=1}^{m_\beta}|f(Y_{\alpha})-f(X_{\beta,i})|^2 \label{eq:LostFunc}
\end{align}
where $Y_\alpha$ denotes the desired image for the $\alpha$-th class, $X_{\alpha,i}$ denotes the $i$-th image in the $\alpha$-th class of the dataset, $f$ denotes the nonlinear mapping represented by the neural network which produces the embedding vector for any given image, as illustrated in Fig.~\ref{fig:DDnet}. In Eq.~(\ref{eq:LostFunc}), $n$ is the number of classes in the dataset, $m_\alpha$ is the number of images belonging to the $\alpha$-th class, and $\lambda$ is a hyperparameter to balance the two terms in the bracket, which is set to be 19 in our calculations. In essence, $d_{\alpha,\beta}$ measures the Euclidean distance between the reference $Y_{\alpha}$ and the images belonging to the $\beta$-th class of the original dataset, in the space where the embedding vectors are defined. Therefore, physically the lost function means that each desired reference is required to resemble the images in the same class to the greatest extent, and at the meanwhile differ from the images in the other classes as much as possible.

\begin{figure}[htbp]
	\centering
	\subfigure[MNIST]{
		\begin{minipage}[t]{0.8\linewidth}
			\centering
			\includegraphics[width=\linewidth]{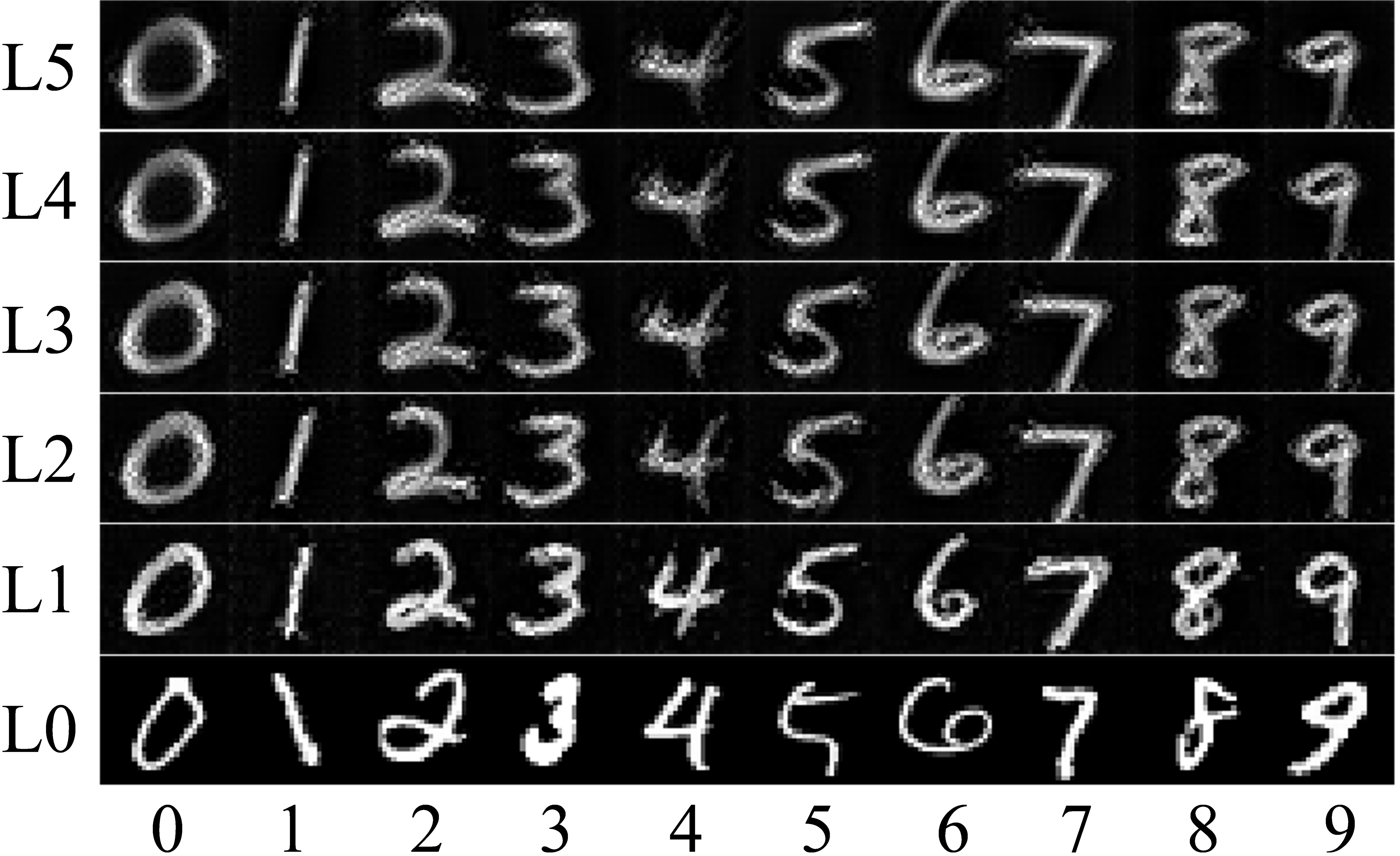}
		\end{minipage}%
	} \\
	\subfigure[CIFAR-10]{
		\begin{minipage}[t]{0.8\linewidth}
			\centering
			\includegraphics[width=\linewidth]{Dcifar.pdf}
		\end{minipage}%
	}%
	\caption{Distilled figures at each abstraction level. For comparison, one sample of each class in the original dataset is shown in $L0$.}
	\label{fig:DataDist}
\end{figure}

The distilled images at different levels are sketched in Fig.~\ref{fig:DataDist}(a). As expected, it seems that the obtained images obtained from lower-level distillations contain more details and are clearer. When the level of distillation goes up, the details gradually blur and only some indescribable features remain. This tendency is more evident for CIFAR-10 dataset, as is shown in Fig.~\ref{fig:DataDist}(b). 

It is reasonable to regard the remaining features in the final distilled images as essential ones which characterize, or even define the dataset in the perfect case. To check this, firstly we train two residual networks, i.e., ResNet18 and ResNet50 on MNIST and CIFAR-10 datasets, respectively, through the usual classification task, and then use the trained models to collect the output embedding vectors \cite{GF-Book2016} of both the test and distilled images. Without resorting to neural networks further, the final classification is performed by directly comparing the similarity between the embedding vectors of an test image and that of the distilled images, according to the angles in between, and the image is classified into a category whose distilled reference image has the highest similarity to it. It shows that this direct comparison can already give test accuracies of about $98.7\%$ and $86.9\%$ for MNIST and CIFAR-10, respectively. This confirms that the above distillation process can indeed capture some essential features of the original dataset, from lower-level abstraction to higher-level abstraction gradually, and this reflects exactly the spirit of deep learning. 

The performance can be further improved in several ways. Firstly, in each distillation procedure, the lost function plays an important role and can be designed more smartly. In this work, the distance between two images is defined as the Euclidean distance in the embedded space, and one can use other measures, such as the Arcface loss \cite{Deng-CVPR2019}, which emphasizes the angular separation and is frequently used in facial-recognition tasks. A preliminary experiment utilizing this lost function can produce an accuracy of about 99\% for MNIST, and 89\% for CIFAR-10. Secondly, the partition of the dataset, as well as the choice of the hyperparameter $\lambda$, might affect the result of the whole distillation. An optimal choice should consider the balance between performance and efficiency in a better way, while in this work, we just adopt the most convenient choice.

\section{Summary}
\label{sec:Summary}
To summarize, inspired by the similarity between the RG flow in physical systems and the inference process in deep neural networks, we introduce the hierarchical coarse-grained structure into the artificial neural networks manifestly to improve the interpretability without degrading performance. To be specific, in the first part, we propose the TaylorNet by introducing the CG operation locally and hierarchically, which extends the linear convolution operation to nonlinear polynomial combinations. It approximates the mapping from the input signal to output result by Taylor expansions effectively, without resorting to the activation functions, and achieves satisfying results in the classification experiments on MNIST and CIFAR-10 datasets. In the distillation task, we propose a setup with a hierarchical coarse-grained structure, and make the inference process from lower levels to higher levels more transparent. It seems that the multistep distillation process is able to capture some essential features of the original dataset, and the distilled images possess less irrelevant details and can be used as reference images in classification tasks. In both cases, the resulting processes represented by the neural networks are more understandable, and the performance are very acceptable compared to the conventional neural networks.

Besides the specific issues discussed separately in Sec.~\ref{sec:TaylorNet} and Sec.~\ref{sec:DataDist}, there are some other aspects that can be explored to further improve the performance. For example, we can use more than one TaylorNets to approximate a single mapping, and put the orthogonality constraint on these networks appropriately for higher efficiency. This might be achieved by adding penalties in the lost functions, training in momentum space, or using other orthogonal complete sets like spherical functions. All these topics are interesting and have been discussed in physical systems, but are far beyond the scope of this paper, and we would like to leave them as pursuits in the near future.  

As to the TaylorNet, it is also worth mentioning that our proposal is very different from the previous work in the literature \cite{Chen-IE2009, Muller-PR2017, Novikov-ICLR2017, Tong-JCP2021, SH-arXiv2021}, which have also explored the Taylor series in neural networks. Most of them have specific motivations and work in different frameworks, and there is no explicit hierarchical structure employed there. For example, Chen \tit{et al.} \cite{Chen-IE2009} used a single-layer neural network similar to Eq.~(\ref{eq:NaTay}) to approximate the Taylor expansion of a single-variable function. Montavon \tit{et al.} \cite{Muller-PR2017} explored the role of Taylor coefficients as derivatives in an ordinary neural network to analyze the importance of a single pixel in the classification task. Tong \tit{et al.} \cite{Tong-JCP2021} used the Taylor series to approximate the quadratic form of a Hermitian matrix. Rao \tit{et al.} \cite{SH-arXiv2021} expressed part of the nonlinearity in terms of a direct-product operation and applied it to the study of partial differential equations. The closest one to our work is probably Ref.~\cite{Novikov-ICLR2017}, where Novikov $\tit{et al.}$ used the neural network to approximate the Taylor expansions; however, in their work both the Taylor terms and Taylor coefficients are represented as compact matrix product operators approximately; thus there is no coarse-grained structure emphasized in this work at all.


At last, though the fixed-point is not discussed at all in this work, the introduction of the hierarchical coarse-grained structure does provide the possibility of its existence. It is possible to study the fixed point of the scale transformations introduced in the TaylorNet, as well as the fixed point of the iterative distillation procedure introduced in Sec.~\ref{sec:DataDist}. Whether the scale invariance can be related to some interesting critical phenomena in this context, as explored in Ref.~\cite{Saeed-PNAS2013}, is an open question deserving attention. 

\section*{Acknowledgement}
We thank Jing Zhang for her contribution in the early stage of this work, and thank Tao Xiang and Ze-Feng Gao for helpful discussions. We are supported by the National R$\&$D Program of China (Grants No. 2017YFA0302900 and No. 2016YFA0300503), the National Natural Science Foundation of China (Grants No. 52176064, 12274458 and No. 11774420), and by the Research Funds of Renmin University of China (Grants No. 20XNLG19).
    
\appendix
\begin{appendix}
\section{Complete expression of $z$}
\label{app:Yexp}
\begin{figure*}
	\includegraphics[scale=0.3]{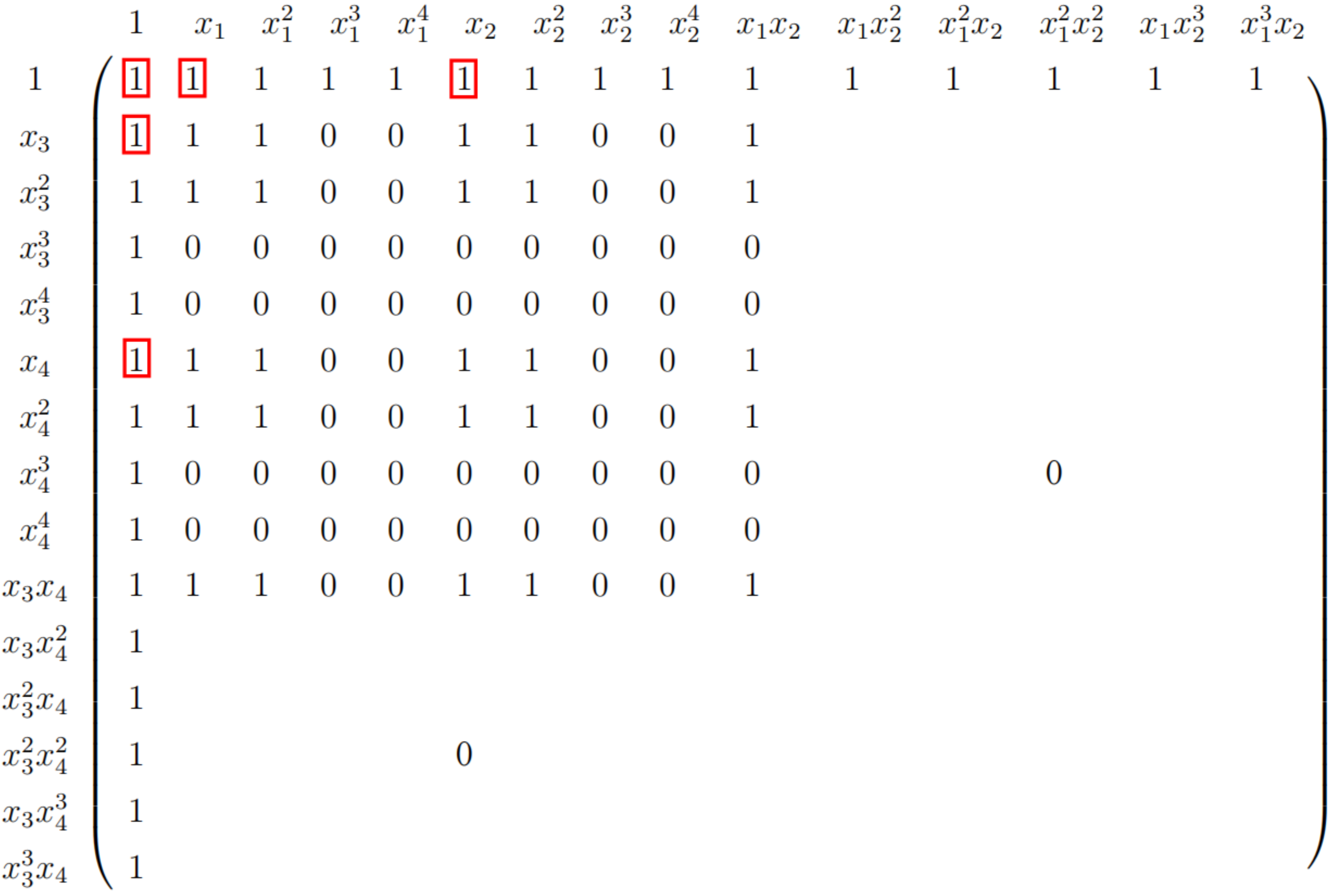}
	\caption{Weight W appearing in Eq.~(\ref{eq:YaWb}), corresponding to the coefficients of $z$ illustrated in Fig.~\ref{fig:CGcore}(c). For simplicity, the nonzero elements are denoted as 1. The terms in the red box are the ones showing up in the convolution operations.}
	\label{fig:W}
\end{figure*}

As illustrated in Fig.~\ref{fig:CGcore}(c), if each local mapping is expanded to the second order, then the complete expression of $z$ in terms of $x_i$, with $i=1,2,3,4$, can be written as
\begin{equation}
	z = \mathbf{a}W\mathbf{b}^{T}, \label{eq:YaWb}
\end{equation}
where the two vectors $\mathbf{a}$ and $\mathbf{b}$ are defined as
\begin{widetext}
		\begin{eqnarray}
			\mathbf{a} & = & \left[\begin{array}{lllllllllllllll}1 & x_{1} & x_{1}^{2} & x_{1}^{3} & x_{1}^{4} & x_{2} & x_{2}^{2} & x_{2}^{3} & x_{2}^{4} & x_{1} x_{2} & x_{1} x_{2}^{2} & x_{1}^{2} x_{2} & x_{1}^{2} x_{2}^{2} & x_{1} x_{2}^{3} & x_{1}^{3} x_{2}\end{array}\right] \nonumber \\
			\mathbf{b} & = & \left[\begin{array}{lllllllllllllll}1 & x_{3} & x_{3}^{2} & x_{3}^{3} & x_{3}^{4} & x_{4} & x_{4}^{2} & x_{4}^{3} & x_{4}^{4} & x_{3} x_{4} & x_{3} x_{4}^{2} & x_{3}^{2} x_{4} & x_{3}^{2} x_{4}^{2} & x_{3} x_{4}^{3} & x_{3}^{3} x_{4}\end{array}\right]. 
		\end{eqnarray}
\end{widetext}
The weight $W$ can be represented as a 15$\times$15 matrix, whose nonzero elements are denoted as 1 in Fig.~\ref{fig:W}, just for clarity.

\section{The action of multi-channel convolutions}
\label{app:Conv}
\begin{figure*}
	\includegraphics[scale=0.08]{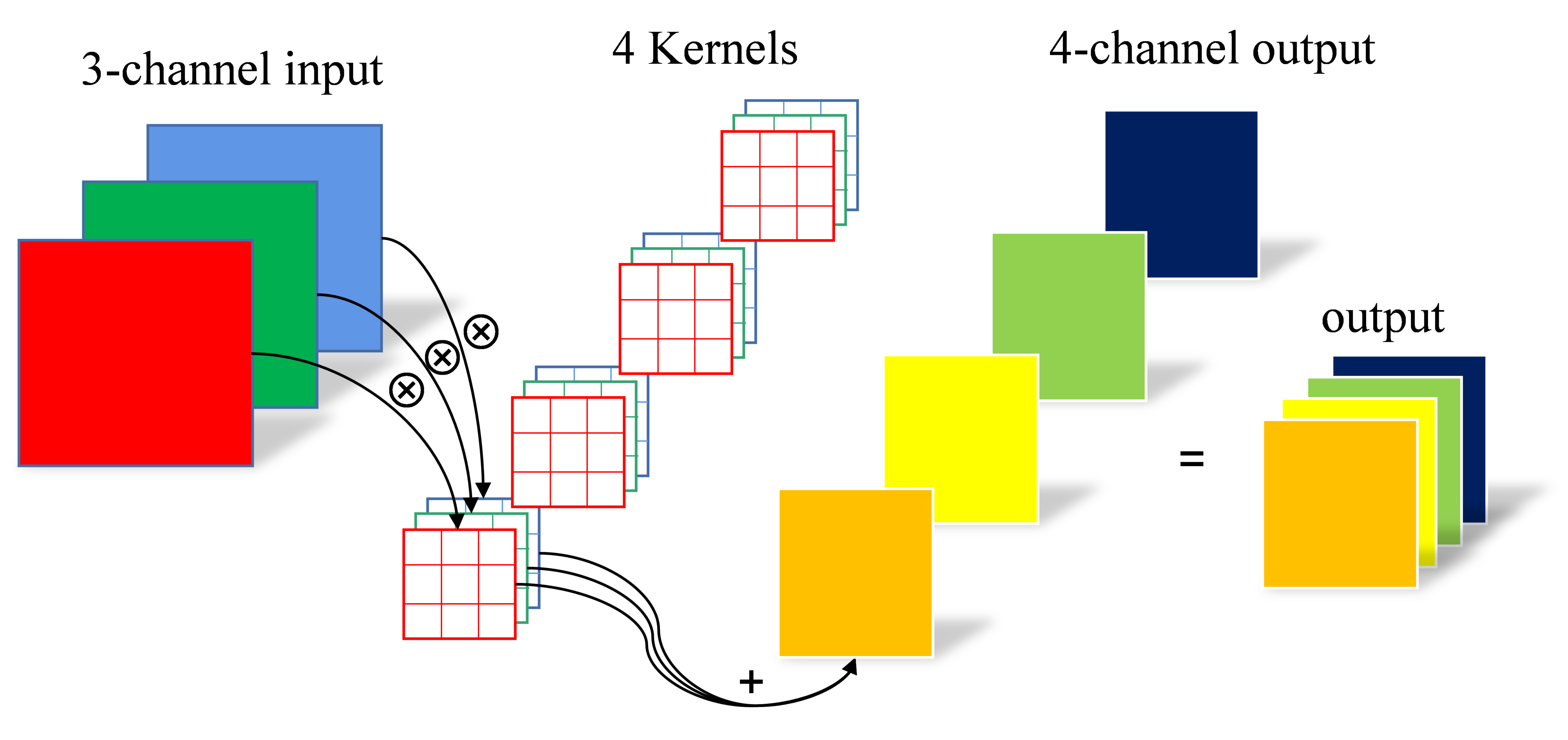}
	\caption{The action of a convolutional layer with four three-channel kernels, as mentioned in Sec.~\ref{sec:TaylorNet}.}
	\label{fig:ConvOP}
\end{figure*}

For concreteness, the convolutional layer with four three-channel kernels will turn the three-channel input data into four-channel output, as illustrated in Fig.~\ref{fig:ConvOP}, where the $+$ sign means equal-weight superposition of the three dot-product results.

\section{TaylorNet structure used in the classification on CIFAR-10 dataset}
\label{app:NetStr}
\begin{figure*}
	\includegraphics[scale=0.06]{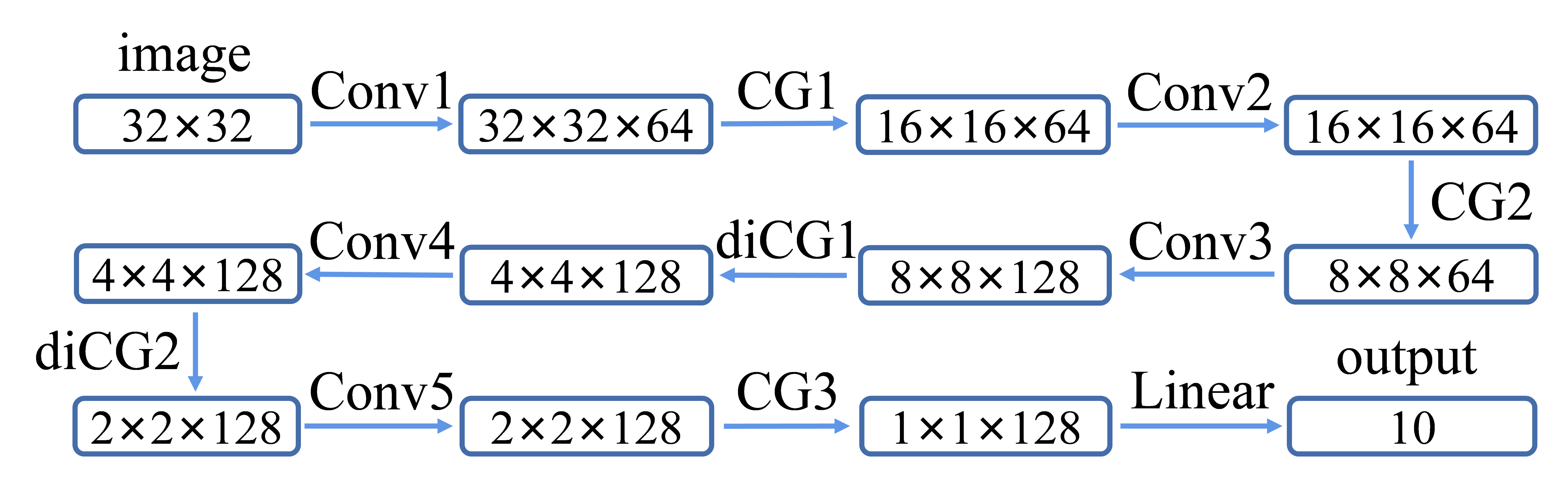}
	\caption{The TaylorNet used in the classification on CIFAR-10 dataset, as mentioned in Sec.~\ref{sec:TaylorNet}. Convolutions Conv1 and Conv2 have structure Conv(3,3,64,1,1,1), Conv3, Conv4, and Conv5 have structure Conv(3,3,128,1,1,1). CG operations CG1 and CG2 have structure CG(2,2,64,2,2), CG3 have structure CG(2,2,128,1,1). Dilated CG operations diCG1 and diCG2 have structures diCG(2,2,128,1,1,4) and diCG(2,2,128,1,1,2), respectively.}
	\label{fig:NetStrCifar}
\end{figure*}

The detailed structure of the TaylorNet used in Sec.~\ref{sec:TaylorNet} in the classification on CIFRA-10 dataset, is shown in Fig.~\ref{fig:NetStrCifar}.

\section{Weight details of the quadratic terms in the experiment on MNIST dataset}
\label{app:Weight}
\begin{figure*}[htbp]
	\centering
	\subfigure[Class 0]{
		\begin{minipage}[t]{0.25\linewidth}
			\centering
			\includegraphics[width=\linewidth]{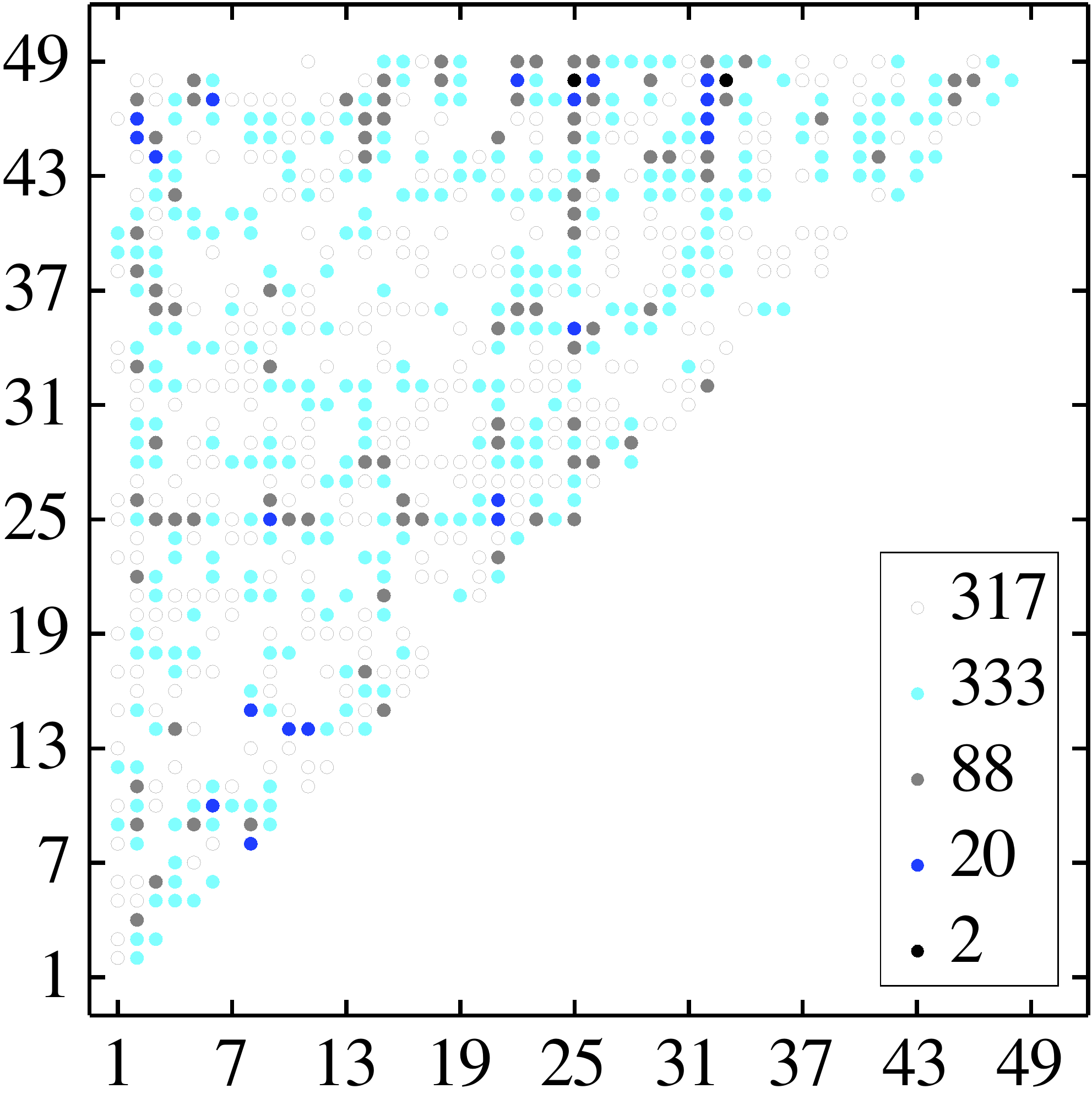}
		\end{minipage}%
	}%
	\subfigure[Class 1]{
		\begin{minipage}[t]{0.25\linewidth}
			\centering
			\includegraphics[width=\linewidth]{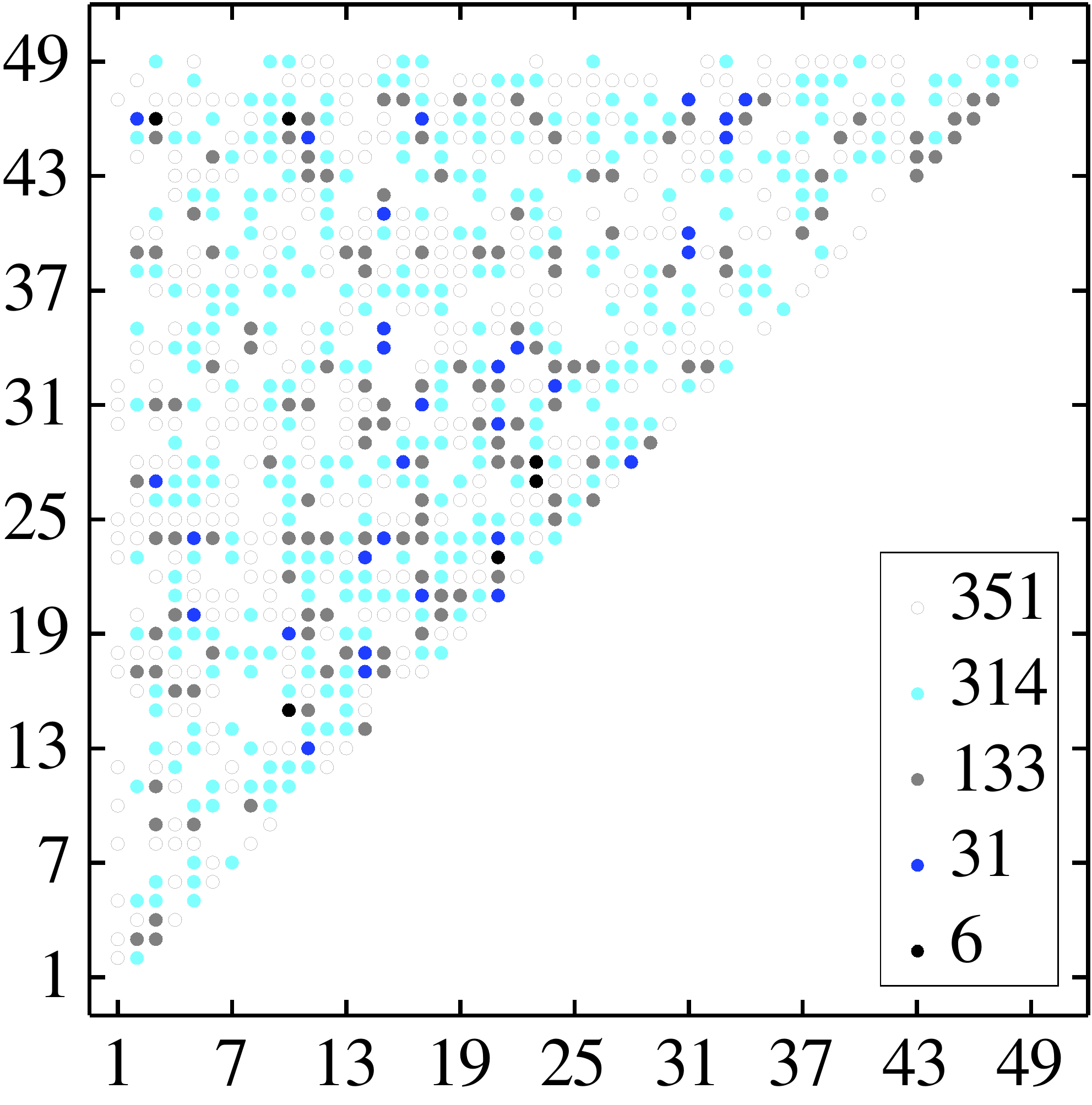}
		\end{minipage}%
	}%
	\subfigure[Class 2]{
		\begin{minipage}[t]{0.25\linewidth}
			\centering
			\includegraphics[width=\linewidth]{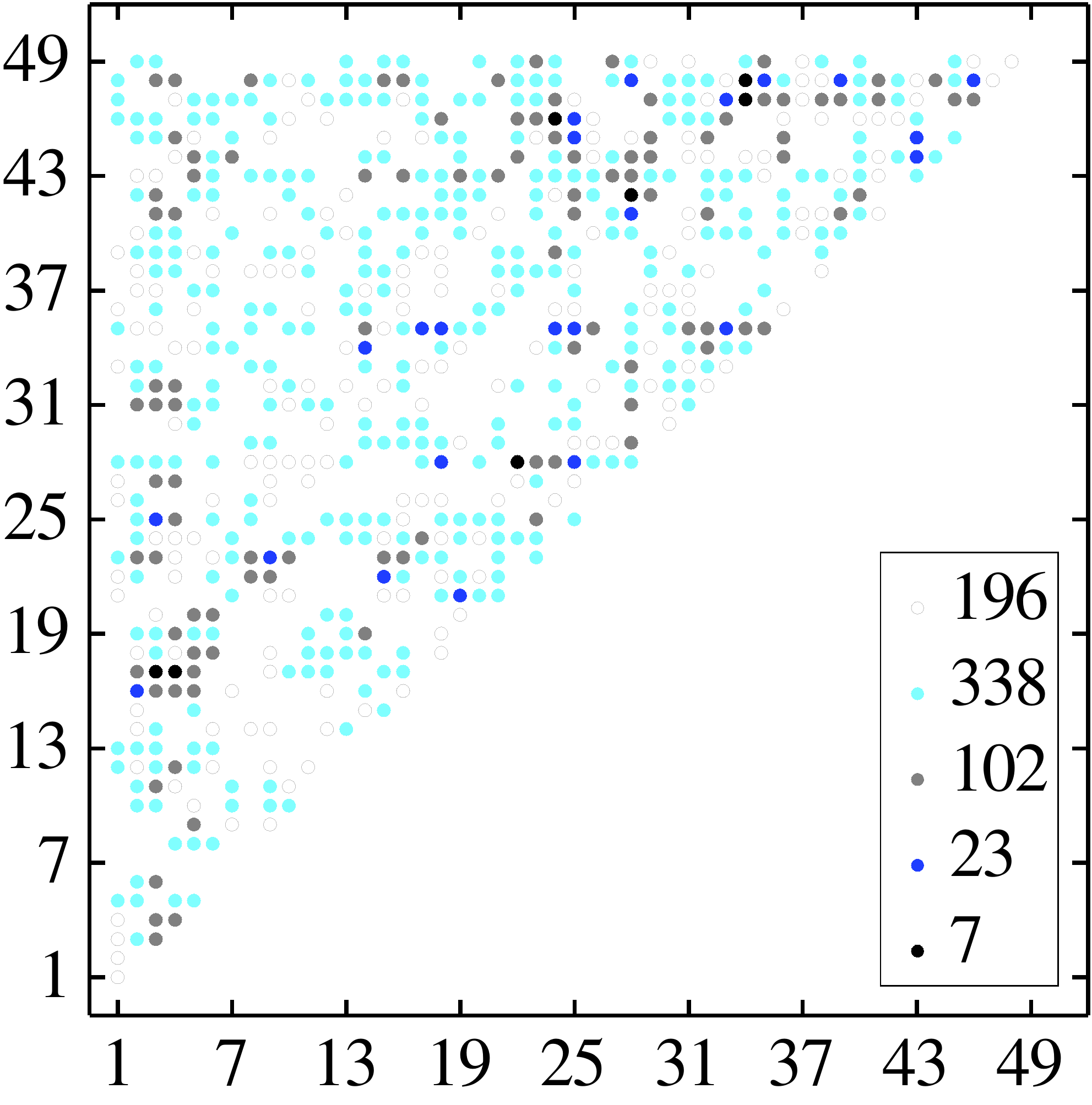}
		\end{minipage}
	}\\
	\subfigure[Class 3]{
		\begin{minipage}[t]{0.25\linewidth}
			\centering
			\includegraphics[width=\linewidth]{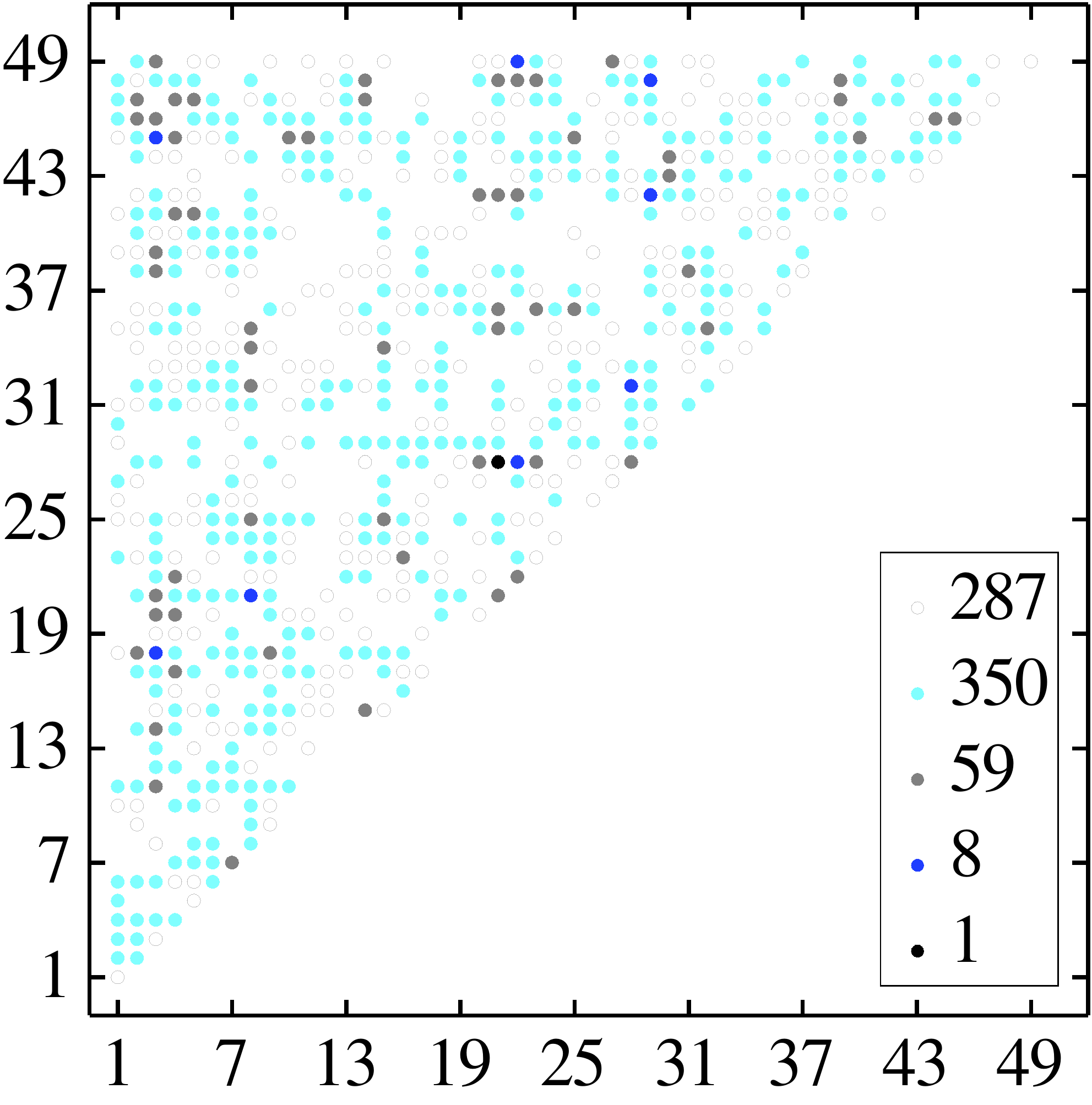}
		\end{minipage}%
	}%
	\subfigure[Class 4]{
		\begin{minipage}[t]{0.25\linewidth}
			\centering
			\includegraphics[width=\linewidth]{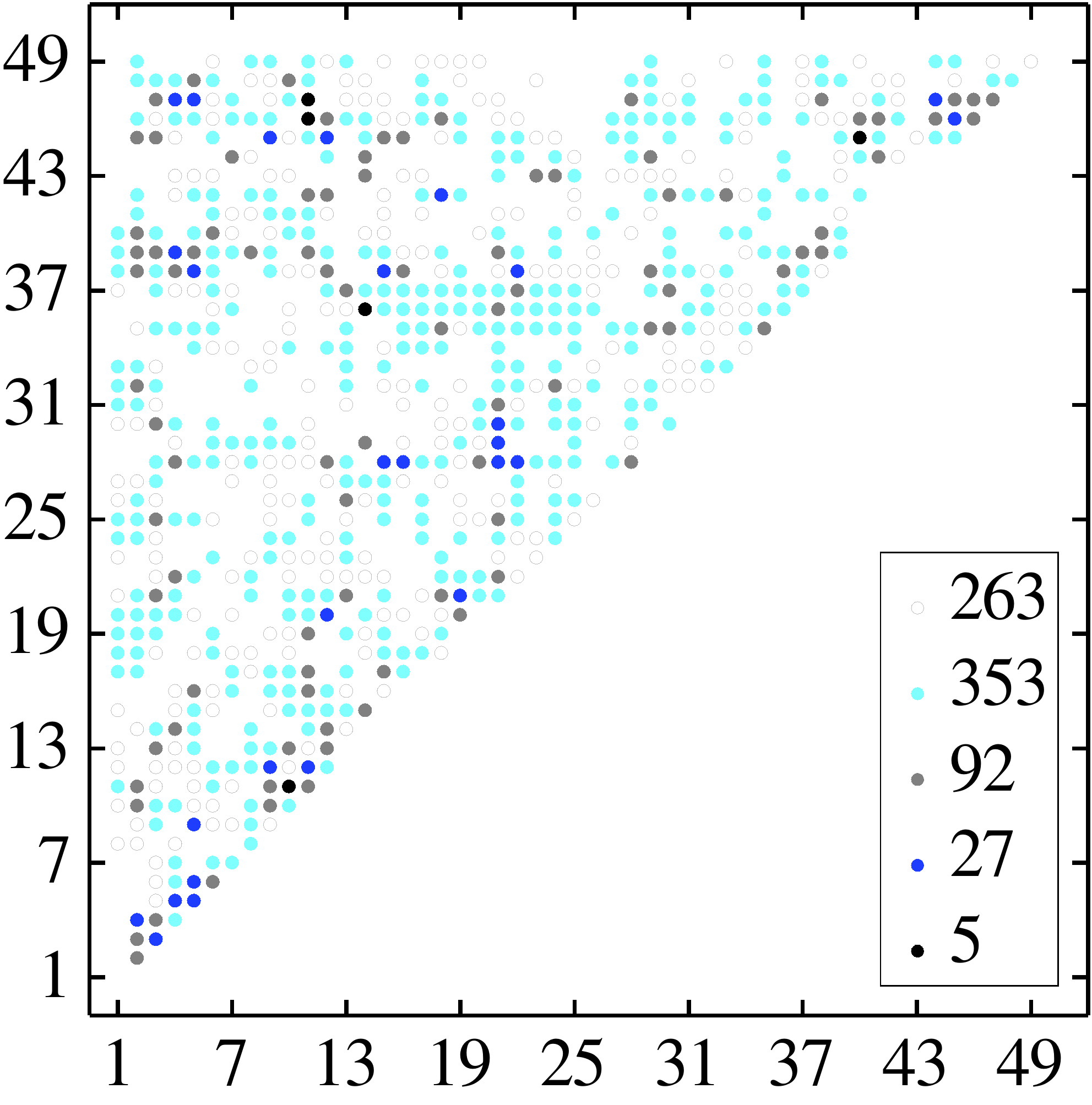}
		\end{minipage}%
	}%
	\subfigure[Class 5]{
		\begin{minipage}[t]{0.25\linewidth}
			\centering
			\includegraphics[width=\linewidth]{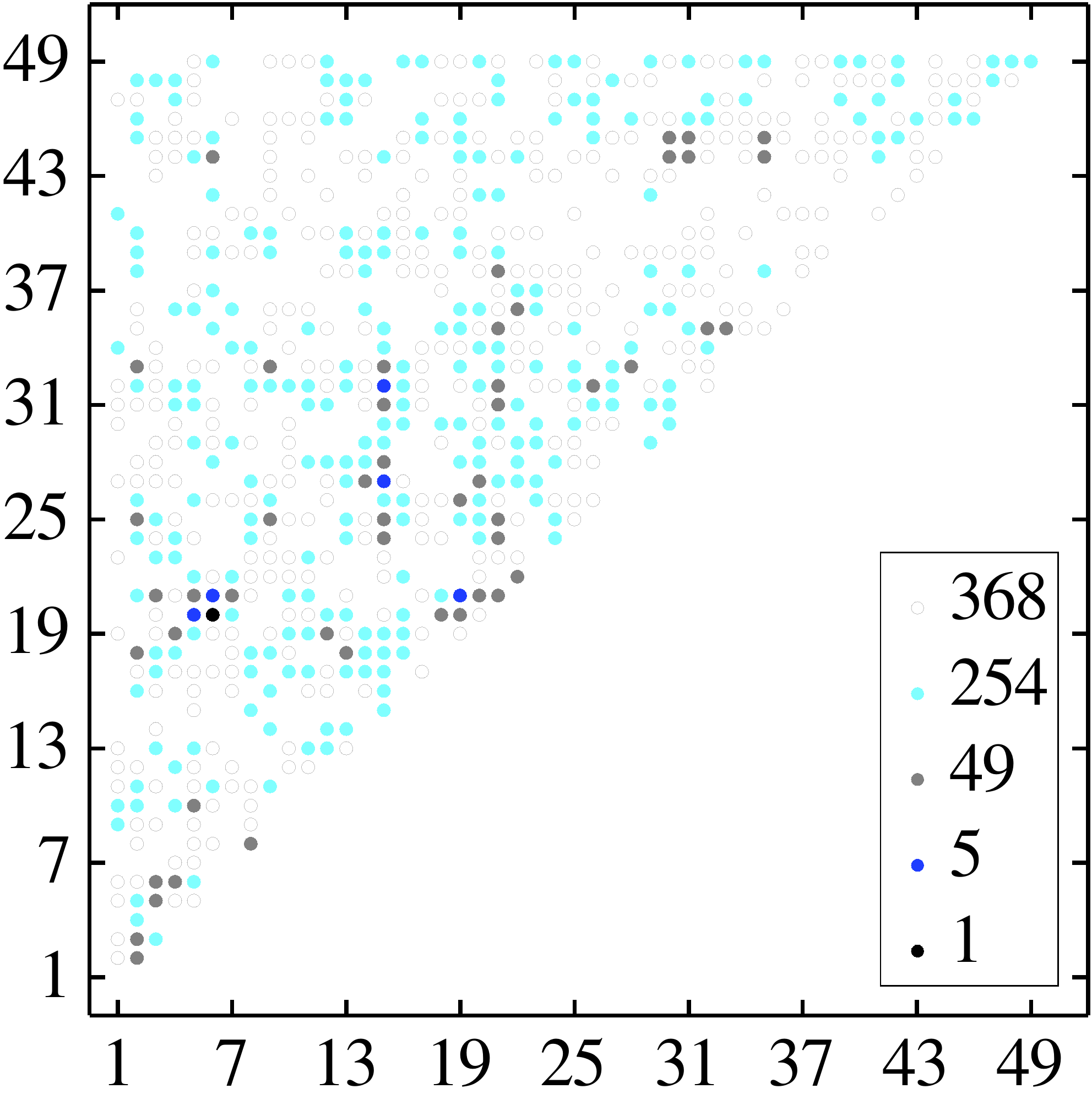}
		\end{minipage}
	}\\
	\subfigure[Class 6]{
		\begin{minipage}[t]{0.25\linewidth}
			\centering
			\includegraphics[width=\linewidth]{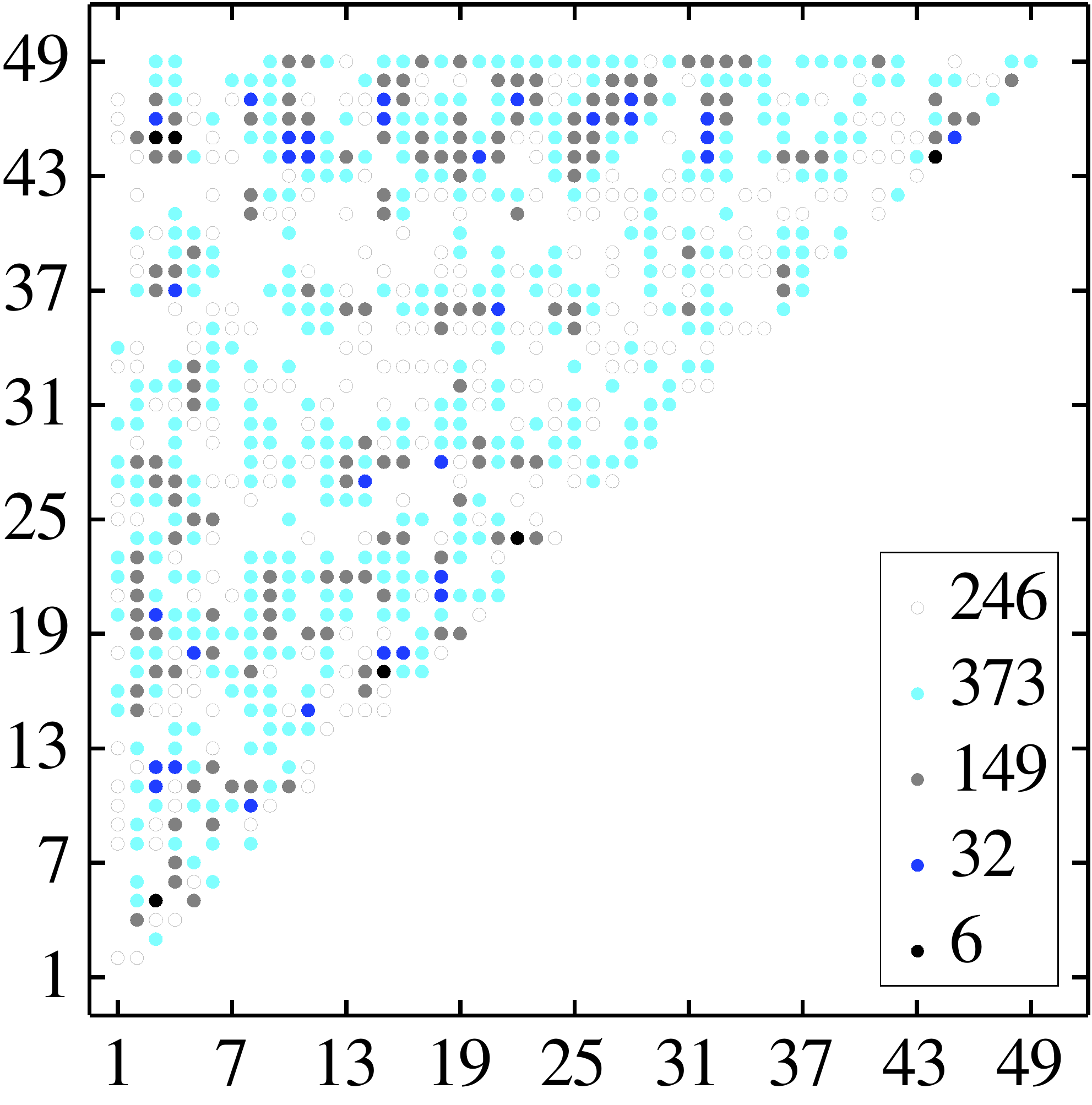}
		\end{minipage}%
	}%
	\subfigure[Class 7]{
		\begin{minipage}[t]{0.25\linewidth}
			\centering
			\includegraphics[width=\linewidth]{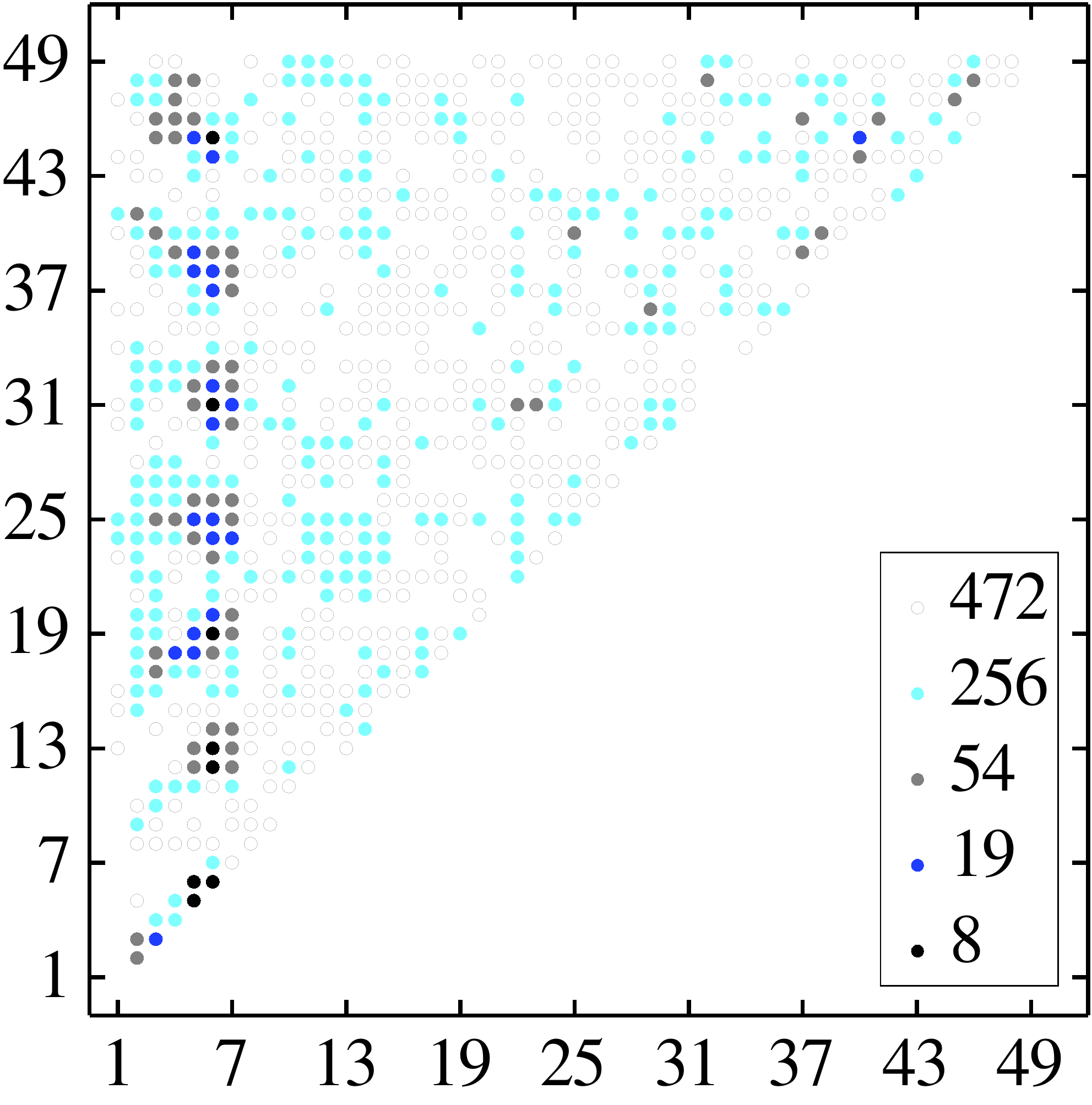}
		\end{minipage}%
	}%
	\subfigure[Class 8]{
		\begin{minipage}[t]{0.25\linewidth}
			\centering
			\includegraphics[width=\linewidth]{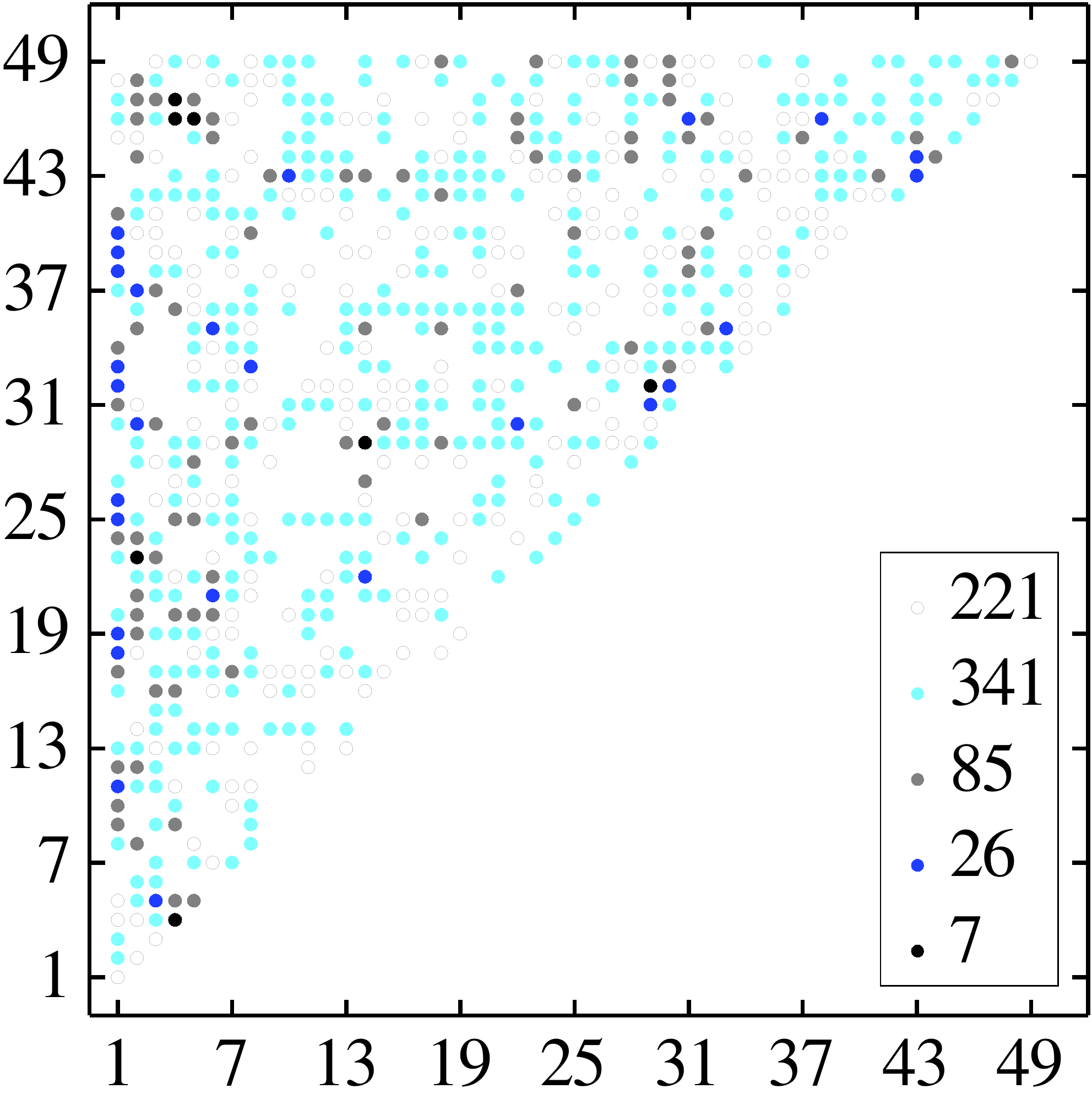}
		\end{minipage}
	}\\
	\subfigure[Class 9]{
		\begin{minipage}[t]{0.25\linewidth}
			\centering
			\includegraphics[width=\linewidth]{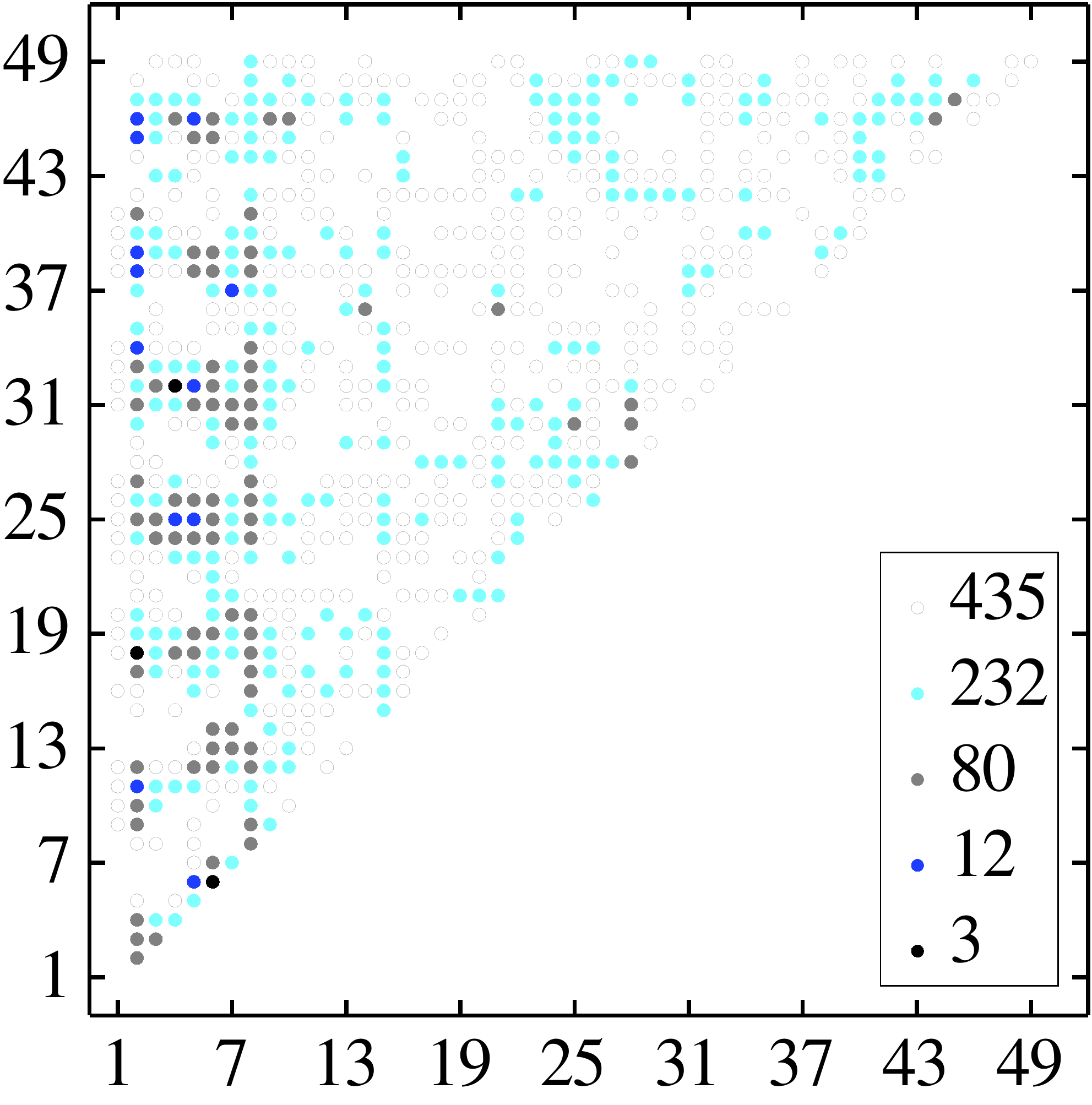}
		\end{minipage}
	}%
	\caption{The weight corresponding to quadratic terms $x_ix_j$ for each class, for experiment of Eq.~(\ref{eq:NaTay}) on the 7$\times$7 MNIST dataset. The abscissa and ordinate represent the linear coordinates $i$ and $j$ in the input image, respectively. The weight is divided into five levels according to the magnitude of the value, with darker colors representing larger weights. The inset counts the number of weights belonging to different levels.}
	\label{fig:DistribDetail}
\end{figure*}

As to the direct experiment on the Taylor expansion, Eq.~(\ref{eq:NaTay}), on MNIST dataset in Sec.~\ref{sec:TaylorNet}, the trained weight of the quadratic terms are sketched in detail in Fig.~\ref{fig:DistribDetail}. The statistics are shown in Fig.~\ref{fig:Distrib}.

\end{appendix}

\end{document}